\pdfoutput=1
\documentclass[11pt]{article}

\usepackage[margin=1in]{geometry}
\usepackage{mathpazo}
\usepackage[backend=biber,style=alphabetic,natbib=true,maxnames=99,maxcitenames=2,minalphanames=3]{biblatex}
\usepackage[colorlinks=true,linkcolor=blue,citecolor=blue,urlcolor=blue]{hyperref}

\usepackage{graphicx}
\usepackage{psfrag}
\usepackage{amsmath}
\usepackage{amsfonts}
\usepackage{verbatim}
\usepackage{optidef}
\usepackage{mathrsfs}
\usepackage{hyperref}
\usepackage{color}
\usepackage[utf8]{inputenc}
\usepackage[T1]{fontenc}

\usepackage{subcaption}
\usepackage{booktabs}
\usepackage{tabularx}
\usepackage{multirow}
\usepackage{float}

\usepackage[ruled,vlined]{algorithm2e}
\SetKwInOut{Input}{Input}
\SetKwInOut{Parameters}{Parameters}

\usepackage[textsize=scriptsize]{todonotes}

\newif\ificml

\icmlfalse

\makeatletter\let\c@table\c@figure\makeatother

\title{Combining Diverse Feature Priors}
\author{
    Saachi Jain\thanks{Equal contribution.} \\
	MIT \\
	\texttt{saachij@mit.edu} \\
	\and
	Dimitris Tsipras\footnotemark[1] \\
	MIT \\
	\texttt{tsipras@mit.edu}
	\and
	Aleksander M\k{a}dry \\
	MIT \\
	\texttt{madry@mit.edu}
}
\date{}

\begin{document}
\maketitle

\begin{abstract}
    \noindent
To improve model generalization, model designers often
restrict the features that their models use, either implicitly or explicitly.
In this work, we explore the design space of leveraging such
\emph{feature priors} by viewing them as distinct perspectives on the data.
Specifically, we find that models trained with diverse sets of 
feature priors have less overlapping failure modes, and can thus be combined
more effectively.
Moreover, we demonstrate that jointly training such models on additional
(unlabeled) data allows them to correct each other's mistakes, which, in turn,
leads to better generalization and resilience to spurious correlations.  
\footnote{Code available at \url{https://github.com/MadryLab/copriors}.}

\end{abstract}

\section{Introduction}
The driving force behind deep learning's success is its ability to automatically
discover predictive features in complex high-dimensional datasets.
These features can generalize beyond the specific task at hand,
thus enabling models to transfer to other (similar) tasks~\citep{donahue2014decaf}.
At the same time, the set of features that the model learns has a large
impact on the model's performance on unseen inputs, especially in the
presence of distribution
shift~\citep{ponce2006dataset,torralba2011unbiased,sagawa2019distributionally} or
spurious correlations~\citep{heinze2017conditional,beery2018recognition,
meinshausen2018causality}.

Motivated by this, recent work focuses on encouraging specific modes of 
 behavior by preventing the models from relying on
certain features.
Examples include suppressing texture
features~\citep{geirhos2018imagenettrained,wang2019learning}, avoiding
$\ell_p$-non-robust features~\citep{tsipras2019robustness,engstrom2019learning}, or utilizing
different parts of the frequency spectrum~\citep{yin2019fourier}.

At a high level, these methods can be thought of as ways of imposing a
\emph{feature prior} on the learning process, so as to bias the model towards
acquiring features that generalize better.
This makes the choice of the feature prior to impose a key design
decision.
The goal of this work is thus to explore the underlying design space of
feature priors and, specifically, to understand:
\begin{center}
  \emph{How can we effectively harness the diversity of feature priors?} 
\end{center}

\subsection*{Our contributions}
In this paper, we cast diverse feature priors as different perspectives on the
data and study how they can complement each other.
In particular, we aim to understand whether training with distinct priors 
result in models with non-overlapping failure modes and how such models can be
combined to improve generalization.
This is particularly relevant in settings where the data is unreliable---e.g, when the training data contains a spurious correlation.
From this perspective, we focus our study on two priors that arise naturally in
the context of image classification, shape and texture, and
investigate the following:

\paragraph{Feature diversity.}
We demonstrate that training models with diverse feature priors results in them making
mistakes on different parts of the data distribution, even if they perform similarly in terms of overall
accuracy.
Further, one can harness this diversity to build model ensembles that
are more accurate than those based on combining models which have the same feature prior.

\paragraph{Combining feature priors on unlabeled data.}
When learning from unlabeled data, the choice of feature prior can be 
especially important. For strategies such as self-training, 
sub-optimal prediction rules learned from sparse labeled data can be 
reinforced when pseudo-labeling the unlabeled data. 
We show that, in such settings, we can leverage the diversity of feature priors to address these 
issues. By \emph{jointly} training models with different feature priors
on the unlabeled data through the framework of \emph{co-training}~\cite{blum1998combining}, we find that the models can correct each other's mistakes to learn prediction rules 
that generalize better. 


\paragraph{Learning in the presence of spurious correlations.}
Finally, we want to understand whether combining diverse priors during training,
as described above, can prevent models from relying on correlations that are
spurious, i.e., correlations that do not hold on the actual distribution of interest.
To model such scenarios, we consider a setting where a spurious correlation is
present in the training data but we also have access to (unlabeled) data where
this correlation does not hold.
In this setting, we find that co-training models with diverse feature priors can
actually steer them away from such correlations and thus enable them to 
generalize to the underlying distribution.\\

Overall, our findings highlight the potential of incorporating
distinct feature priors into the training process.
We believe that further work along this direction will lead us to models that
generalize more reliably.

\section{Background: Feature Priors in Computer Vision}
\label{sec:priors}
\label{sect:priors}
When learning from structurally complex data, such as images, relying on raw input features
alone (e.g., pixels) is not particularly useful.
There has thus been a long line of work on extracting  input patterns
that can be more effective for prediction.
While early approaches, such as
SIFT~\citep{lowe1999object} and HOG~\citep{dalal2005histograms}, leveraged hand-crafted features,
these have been largely replaced by features that are automatically learned in an
end-to-end fashion~\citep{krizhevsky2009learning,ciregan2012multi,krizhevsky2012imagenet}.

Nevertheless, even when features are learned, model designers still tune their models to better suit a
particular task via changes in the architecture or training methodology.
Such modifications can be thought of as imposing \emph{feature priors}, i.e., priors
that bias a model towards a particular set of features.
One prominent example is convolutional neural networks, which are
biased towards learning a hierarchy of localized
features~\cite{fukushima1980neocognitron,lecun1989backpropagation}.
Indeed, such a convolutional prior can be quite powerful: it is sufficient to
enable many image synthesis tasks \emph{without any
training}~\cite{ulyanov2017deep}.

More recently, there has been work exploring the impact of explicitly
restricting the set of features utilized by the model.
For instance, \citet{geirhos2018imagenettrained} demonstrate that training
models on stylized inputs (and hence suppressing texture information) can
improve model robustness to common corruptions.
In a similar vein, \citet{wang2019learning} penalize the predictive power of
local features to learn shape-biased models that generalize better between image
styles.
A parallel line of work focuses on training models to be robust to small,
worst-case input perturbations using, for example, adversarial
training~\cite{goodfellow2015explaining,madry2018towards} or randomized
smoothing~\citep{lecuyer2018certified,cohen2019certified}.
Such methods bias these models away from non-robust
features~\citep{tsipras2019robustness,ilyas2019adversarial,engstrom2019learning}, 
which tends to result in them being more aligned with human
perception~\citep{tsipras2019robustness,kaur2019perceptually}, more resilient to certain input
corruptions~\citep{ford2019adversarial,kireev2021effectiveness}, and better
suited for transfer to downstream
tasks~\cite{utrera2020adversarially,salman2020adversarially}.

\section{Feature Priors as Different Perspectives}
As we discussed, the choice of feature prior can have
a large effect on what features a model relies on and, by extension, on how
well it generalizes to unseen inputs.
In fact, one can view such priors as distinct perspectives on the data, capturing
different information about the input.
In this section, we provide evidence to support this view; specifically, we examine a case
study on a pair of feature priors that arise naturally in the context of image
classification: shape and texture.

\subsection{Training shape- and texture-biased models}
\label{sect:shape_texture}
\begin{figure}[!h]
    \centering
    \hfill
    \begin{subfigure}[b]{0.15\textwidth}
        \centering
        \includegraphics[width=\linewidth]{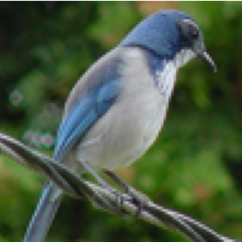}
        \caption{Original}
        \label{fig:prior_original}
    \end{subfigure}
    \hfill
    \begin{subfigure}[b]{0.15\textwidth}
        \centering
        \includegraphics[width=\linewidth]{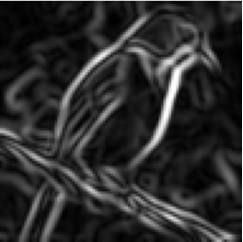}
        \caption{Sobel}
        \label{fig:prior_sobel}
    \end{subfigure}
    \hfill
\ificml
    \phantom{}
    \\
    \hfill
\else
\fi
    \begin{subfigure}[b]{0.15\textwidth}
        \centering
        \includegraphics[width=\linewidth]{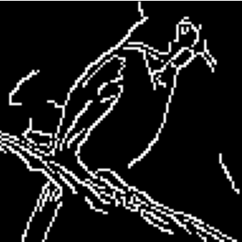}
        \caption{Canny}
        \label{fig:prior_canny}
    \end{subfigure}
    \hfill
    \begin{subfigure}[b]{0.15\textwidth}
        \centering
        \includegraphics[width=\linewidth]{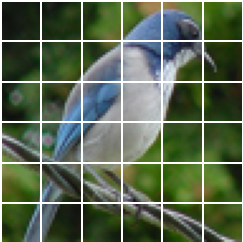}
        \caption{BagNet}
        \label{fig:prior_bagnet}
    \end{subfigure}
    \hfill
    \phantom{}

    \caption{Visualizing different feature priors: (a) an image from the STL-10
        dataset; (b) Sobel edge detection; (c) Canny
    edge detection; (d) the limited receptive field of a BagNet.}
    \ificml
    \vspace{-1em}
    \else
    \fi
    \label{fig:priors}
\end{figure}

In order to train shape- and texture-biased models, we either pre-process the model
input or modify the model architecture as follows:

\paragraph{Shape-biased models.} 
To suppress texture information, we pre-process our
images by applying an edge detection algorithm.
We consider two canonical methods: the \emph{Canny} algorithm~\cite{ding2001canny} which
produces a binary edge mask, and the \emph{Sobel} 
algorithm~\cite{sobel19683x3} which provides a softer edge detection, hence retaining
some texture information (see Figures~\ref{fig:prior_sobel} and~\ref{fig:prior_canny}).

\paragraph{Texture-biased models.}
To prevent the model from relying on the global structure of the
image, we utilize a variant of the \emph{BagNet}
architecture~\cite{brendel2018approximating}.
A BagNet deliberately limits the model's receptive field, thus forcing it to rely on local features
(see Figure~\ref{fig:prior_bagnet}).

\noindent
We visualize all of these priors in Figure~\ref{fig:priors} and provide 
implementation details in Appendix~\ref{app:setup}.

\begin{table*}[!h]
    \centering
        \ificml
        \else
            \setlength{\tabcolsep}{.7em}
\def\arraystretch{1.1}
\begin{tabular}{l|cccc|cccc}
        \toprule
        & \multicolumn{4}{c|}{CIFAR-10}
        & \multicolumn{4}{c}{STL-10} \\
                 & Standard & Canny & Sobel & BagNet & Standard & Canny & Sobel & BagNet \\
        \midrule
        Standard &    0.598 & 0.237 & 0.259 &   0.38 &    0.554 & 0.305 & 0.385 &  0.357 \\
        Canny &          & 0.545 & 0.324 &  \textbf{0.143} &          & 0.523 &
        0.392 &  \textbf{0.212} \\
           Sobel &          &       & 0.594 &  0.173 &          &       & 0.649 &  0.262 \\
          BagNet &          &       &       &  0.655 &          &       &       &  0.486 \\
        \bottomrule
\end{tabular}
%







        \fi
        \caption{Correlation (Pearson coefficient) of correct predictions on the
        test set between different pairs of models.
        The diagonal entries correspond to models trained with the same prior
        but from different random initializations.
        While the two shape-biased models (Sobel and Canny) are
        more aligned with each other, they are both quite different from the
        texture-biased model (BagNet). }
        \phantom{}
        \ificml
            
        \else
        \fi
    \label{tab:independence}
\end{table*}

\subsection{Diversity of feature-biased models}
After training models with shape and texture biases as outlined above, we
evaluate whether these models indeed capture complementary information about
the input.
Specifically, we train models on a small subset (100 examples per class)
of the CIFAR-10~\citep{krizhevsky2009learning} and
STL-10~\citep{coates2011analysis} datasets, and measure the correlation
between which test examples they correctly classify. We consider the full CIFAR-10 dataset, 
as well as similar experiments on ImageNet~\cite{deng2009imagenet} in Appendix~\ref{app:full_cifar10_IN}.

We find that pairs consisting of a shape-biased model and a texture-biased model
(i.e., Canny and BagNet, or Sobel and BagNet) indeed have the least correlated predictions---cf. Table~\ref{tab:independence}.
In other words, the mistakes that these models make are more
diverse than those made by identical models trained from different random
initializations.
At the same time, different shape-biased models (Sobel and Canny) are relatively
well-correlated with each other, which corroborates the fact that models trained on
similar features of the input are likely to make similar mistakes.

\paragraph{Model ensembles.}
\label{ensembles} Having shown that training models with these feature priors 
results in diverse prediction rules, we examine if we can now combine them to improve our
generalization. 
The canonical approach for doing so is to incorporate these models into an
ensemble. 

We find that the diversity of models trained with different feature priors directly translates 
into improved performance when combining them into an
ensemble---cf.\ Table~\ref{tab:ensembles}.
Indeed, we find that the ensemble's performance is tightly connected to the
prediction similarity of its constituents (as measured in
Table~\ref{tab:independence}), i.e., more diverse ensembles tend to perform better.
For instance, the best ensemble for the STL-10 dataset combines a
shape-biased model (Canny) and a texture-biased model (BagNet) which were the
models with the least aligned predictions.

\begin{table*}[!h]
    \ificml
    \else
        \begin{subfigure}[b]{\linewidth}
            \begin{center}
                \setlength{\tabcolsep}{.7em}
\def\arraystretch{1.1}
\begin{tabular}{llccc}
    \toprule
                                     &      Feature Priors &          Model 1 &          Model 2 &         Ensemble \\
    \midrule
               \multirow{3}{*}{Same} & Standard + Standard & 52.54 $\pm$ 0.86 & 51.82 $\pm$ 0.86 & 54.02 $\pm$ 0.80 \\
                                     &       Sobel + Sobel & 51.94 $\pm$ 0.84 & 53.69 $\pm$ 0.82 & 54.68 $\pm$ 0.83 \\
                                     &     BagNet + BagNet & 42.22 $\pm$ 0.88 & 42.56 $\pm$ 0.80 & 43.49 $\pm$ 0.83 \\
    \hline\multirow{3}{*}{Different} &    Standard + Sobel & 52.54 $\pm$ 0.83 & 51.94 $\pm$ 0.83 & \textbf{58.21 $\pm$ 0.82} \\
                                     &   Standard + BagNet & 52.54 $\pm$ 0.84 & 42.22 $\pm$ 0.84 & 53.03 $\pm$ 0.81 \\
                                     &      Sobel + BagNet & 51.94 $\pm$ 0.90 & 42.22 $\pm$ 0.84 & 55.14 $\pm$ 0.81 \\
    \bottomrule
\end{tabular}


            \end{center}
            \vspace{-1em}
            \caption{CIFAR-10}
            \vspace{1em}
        \end{subfigure}
        \begin{subfigure}[b]{\linewidth}
            \begin{center}
                \setlength{\tabcolsep}{.7em}
\def\arraystretch{1.1}
\begin{tabular}{llccc}
    \toprule
                                     &      Feature Priors &          Model 1 &          Model 2 &         Ensemble \\
    \midrule
               \multirow{3}{*}{Same} & Standard + Standard & 53.73 $\pm$ 0.91 & 55.38 $\pm$ 0.88 & 57.06 $\pm$ 0.91 \\
                                     &       Canny + Canny & 56.29 $\pm$ 0.96 & 54.99 $\pm$ 0.96 & 58.23 $\pm$ 0.93 \\
                                     &     BagNet + BagNet & 52.04 $\pm$ 0.98 & 50.34 $\pm$ 0.94 & 53.42 $\pm$ 0.93 \\
    \hline\multirow{3}{*}{Different} &    Standard + Canny & 53.73 $\pm$ 0.95 &
    56.29 $\pm$ 0.91 & \textbf{60.96 $\pm$ 0.96} \\
                                     &   Standard + BagNet & 53.73 $\pm$ 0.98 & 52.04 $\pm$ 0.90 & 57.17 $\pm$ 0.90 \\
                                     &      Canny + BagNet & 56.29 $\pm$ 0.91 & 52.04 $\pm$ 0.95 & \textbf{61.42 $\pm$ 0.92} \\
    \bottomrule
\end{tabular}



            \end{center}
            \vspace{-1em}
            \caption{STL-10}
        \end{subfigure}
    \fi
    \caption{Ensemble accuracy when combining models trained with a diverse
        set of feature priors (models with the same prior are trained from
        different random initialization).
        Notice how models trained with different priors lead to ensembles with
        better performance.
        Moreover, when the accuracy of the two base models is comparable, models
        that are more diverse (as measured in Table~\ref{tab:independence})
        result in better ensembles.
        We describe the different methods of combining models in
        Appendix~\ref{app:ensemble} and provide the full results in Appendix~\ref{app:full_pretrained_ensemble}.}
    \ificml
        \begin{subfigure}[b]{\linewidth}
            \begin{center}
                \setlength{\tabcolsep}{.7em}
\def\arraystretch{1.1}
\begin{tabular}{llccc}
    \toprule
                                     &      Feature Priors &          Model 1 &          Model 2 &         Ensemble \\
    \midrule
               \multirow{3}{*}{Same} & Standard + Standard & 52.54 $\pm$ 0.86 & 51.82 $\pm$ 0.86 & 54.02 $\pm$ 0.80 \\
                                     &       Sobel + Sobel & 51.94 $\pm$ 0.84 & 53.69 $\pm$ 0.82 & 54.68 $\pm$ 0.83 \\
                                     &     BagNet + BagNet & 42.22 $\pm$ 0.88 & 42.56 $\pm$ 0.80 & 43.49 $\pm$ 0.83 \\
    \hline\multirow{3}{*}{Different} &    Standard + Sobel & 52.54 $\pm$ 0.83 & 51.94 $\pm$ 0.83 & \textbf{58.21 $\pm$ 0.82} \\
                                     &   Standard + BagNet & 52.54 $\pm$ 0.84 & 42.22 $\pm$ 0.84 & 53.03 $\pm$ 0.81 \\
                                     &      Sobel + BagNet & 51.94 $\pm$ 0.90 & 42.22 $\pm$ 0.84 & 55.14 $\pm$ 0.81 \\
    \bottomrule
\end{tabular}


            \end{center}
            \caption{CIFAR-10}
        \end{subfigure}
        \begin{subfigure}[b]{\linewidth}
            \begin{center}
                \setlength{\tabcolsep}{.7em}
\def\arraystretch{1.1}
\begin{tabular}{llccc}
    \toprule
                                     &      Feature Priors &          Model 1 &          Model 2 &         Ensemble \\
    \midrule
               \multirow{3}{*}{Same} & Standard + Standard & 53.73 $\pm$ 0.91 & 55.38 $\pm$ 0.88 & 57.06 $\pm$ 0.91 \\
                                     &       Canny + Canny & 56.29 $\pm$ 0.96 & 54.99 $\pm$ 0.96 & 58.23 $\pm$ 0.93 \\
                                     &     BagNet + BagNet & 52.04 $\pm$ 0.98 & 50.34 $\pm$ 0.94 & 53.42 $\pm$ 0.93 \\
    \hline\multirow{3}{*}{Different} &    Standard + Canny & 53.73 $\pm$ 0.95 &
    56.29 $\pm$ 0.91 & \textbf{60.96 $\pm$ 0.96} \\
                                     &   Standard + BagNet & 53.73 $\pm$ 0.98 & 52.04 $\pm$ 0.90 & 57.17 $\pm$ 0.90 \\
                                     &      Canny + BagNet & 56.29 $\pm$ 0.91 & 52.04 $\pm$ 0.95 & \textbf{61.42 $\pm$ 0.92} \\
    \bottomrule
\end{tabular}



            \end{center}
            \caption{STL-10}
        \end{subfigure}
        \vspace{-1.5em}
    \else
    \fi
    \label{tab:ensembles}
\end{table*}

\section{Combining Diverse Priors on Unlabeled Data}
\label{sect:methodology}
In the previous section, we saw that training models with different
feature priors (e.g., shape- and texture-biased models) can lead to prediction
rules with less overlapping failure modes---which, in turn, can lead to more
effective model ensembles.
However, ensembles only combine model predictions post hoc and thus cannot take
advantage of diversity during the training process.

In this section, we instead focus on utilizing diversity \emph{during} training.
In particular, we leverage the diversity introduced through feature priors in the
context of self-training~\cite{lee2013pseudo}---a framework commonly used when
the labeled data is insufficient to learn a well-generalizing model.
This framework utilizes unlabeled data, which are then
pseudo-labeled using an existing model and used for further training.
While such methods can often improve the overall model performance, they face a significant drawback: models tend to reinforce suboptimal prediction
rules even when these rules do not generalize to the underlying
distribution~\cite{arazo2019pseudo}.

\begin{figure*}[t]
    \hfill
    \begin{subfigure}[b]{0.39\linewidth}
        \centering
        \includegraphics[width=\textwidth]{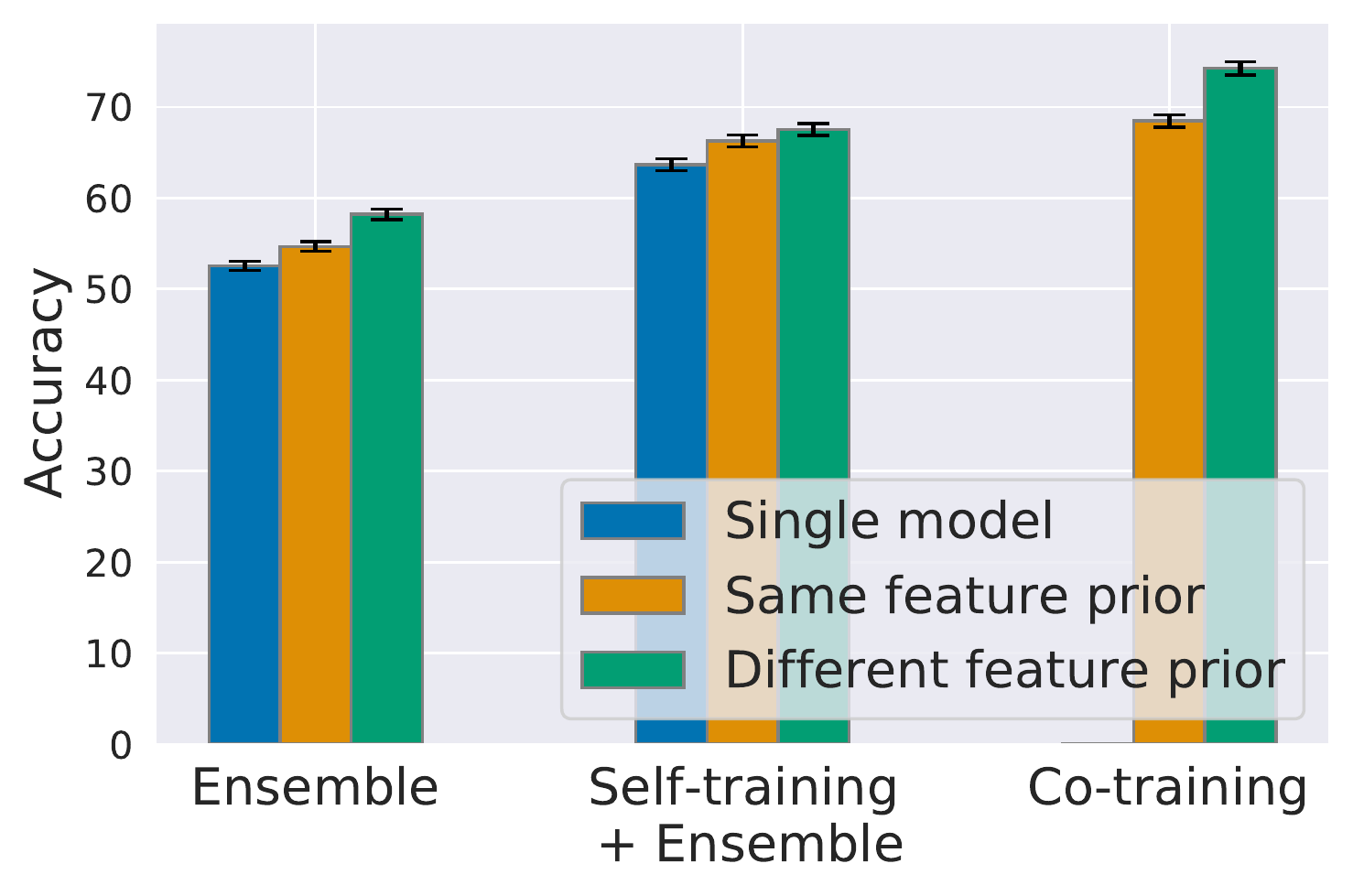}
        \caption{CIFAR-10}
    \end{subfigure}
    \hfill
    \begin{subfigure}[b]{0.39\linewidth}
        \centering
        \includegraphics[width=\textwidth]{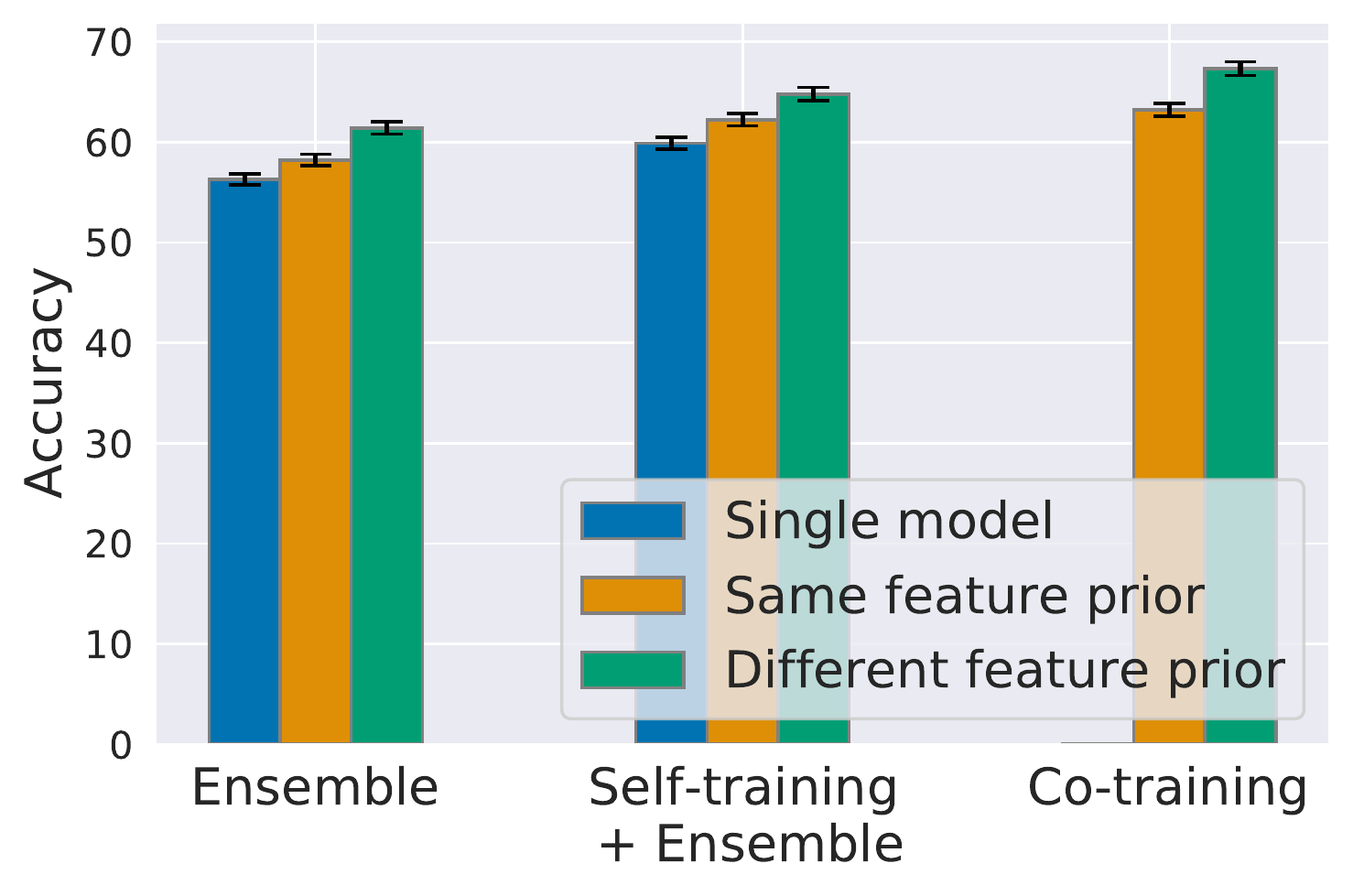}
        \caption{STL-10}
    \end{subfigure}
    \hfill
    \phantom{}
    \caption{Test accuracy of pre-trained, self-trained, and co-trained
        models selecting the best feature prior for each (full results in
        Table~\ref{tab:ensembles}, Appendix
        Table~\ref{tab:self_trained_ensembles}, and 
        Table~\ref{tab:basictable} respectively).
        Notice how combinations of models with different feature priors
        consistently outperform combinations of models with the same 
        feature prior.
    }
    \label{fig:summary}
\end{figure*}

Our goal is thus to leverage diverse feature priors to address this exact
shortcoming.
Specifically, we will \emph{jointly} train models with different priors on the
unlabeled data through the framework of co-training~\cite{blum1998combining}.
Since these models capture complementary information about the input (cf.
Table~\ref{tab:independence}), we expect them to correct each other's mistakes
and improve their prediction rules.
As we will see in this section, this approach can indeed have a significant
impact on the performance of the resulting model, outperforming ensembles that
combine such models only at evaluation time---see summary in Figure~\ref{fig:summary}.

\paragraph{Setup.} We base our analysis on the CIFAR-10 and STL-10 datasets.
Specifically, we treat a small fraction of the training set as \emph{labeled}
examples (100 examples per class), another fraction as our validation set
for tuning hyperparameters (10\% of the total training examples), and the 
rest as \emph{unlabeled} data. We report our results
on the standard test set of each dataset. (See Appendix~\ref{app:setup} for experimental details, and Appendix~\ref{app:vary_label_data_stl10} 
for experiments with varying levels of labeled data.)

\subsection{Self-training and ensembles}
\label{sect:single_prior}
Before outlining our method for jointly training models with multiple
priors, we first describe the standard approach to
self-training a single model.
At a high level, the predictions of the model on the
unlabeled data are treated as correct labels and are then used to further train the same
model~\cite{lee2013pseudo,iscen2019label,zou2019confidence,xie2020self}.
The underlying intuition is that the classifier will predict the correct labels
for that data better than chance, and thus these \emph{pseudo-labels} can be
used to expand the training set. 

In practice, however, these pseudo-labels tend to be noisy. 
Thus, a common approach is to only use the labels to which the model
assigns the highest probability~\cite{lee2013pseudo}.
This process is repeated, self-training on increasingly larger fractions of the
unlabeled data until all of it is used. We refer to each training phase as
an \emph{era}.

\paragraph{Ensembles of diverse self-trained models.}
Similarly to our results in Table~\ref{tab:ensembles}, we find that ensembles
comprised of self-trained models with diverse feature priors outperform those
using same prior from different random initializations
(see~Figure~\ref{fig:summary} for a summary and Appendix~\ref{app:ensemble_self_train}
 for the full results).
This indicates that, after self-training, these models continue to capture
complementary information about the input that can be leveraged to improve
performance.

\subsection{Co-training models with different feature priors}
\label{sec:cotraining}
Moving beyond self-training with a single feature prior, our goal in this
section is to leverage multiple feature priors by jointly training them on 
unlabeled data.
This idea naturally fits into the framework of \emph{co-training}: a method used
to learn from unlabeled data when inputs correspond to multiple
independent sets of features~\cite{blum1998combining}.

Concretely, we first train a model for each feature prior.
Then, we collect the pseudo-labels on the unlabeled data that were assigned 
the highest probability for each
model---including duplicates with potentially different labels---to form a new
training set which we use for further training.
As in the self-training regime, we repeat this process over several eras,
increasing the fraction of the unlabeled dataset used at each era.
Intuitively, this iterative process allows the models to bootstrap off of each
other's predictions, learning correlations that they might fail to learn from
the labeled data alone.
At the end of this process, we are left with two models, one for each prior, which we combine into a single classifier by training a
standard model from scratch on the combined pseudo-labels.
We provide a more detailed explanation of the methodology in
Appendix~\ref{app:co_training_algos}.

\begin{table*}[!ht]
    \ificml 
    \else
        \begin{subfigure}[b]{\linewidth}
        \begin{center}
            \begin{tabular}{llccc}
      \toprule
                                    Methods &  Prior(s) &     Labeled Only & \begin{tabular}{c}+Unlabeled \\Self/Co-Training\end{tabular} & \begin{tabular}{c} + Standard model \\with Pseudo-labels \end{tabular} \\
      \midrule
             \multirow{3}{*}{Self-training} &  Standard & 52.54 $\pm$ 0.81 &                                   63.65 $\pm$ 0.78 &                                   64.02 $\pm$ 0.79 \\
                                            &     Sobel & 51.94 $\pm$ 0.90 &                                   63.05 $\pm$ 0.85 &                                   64.77 $\pm$ 0.81 \\
                                            &    BagNet & 42.22 $\pm$ 0.81 &                                   53.92 $\pm$ 0.84 &                                   54.21 $\pm$ 0.81 \\
      \midrule \multirow{4}{*}{Co-training} &  Standard & 52.54 $\pm$ 0.83 &                                   65.06 $\pm$ 0.78 &                  \multirow{2}{*}{65.10 $\pm$ 0.79} \\
                                            & +Standard & 51.82 $\pm$ 0.79 &                                   64.93 $\pm$ 0.83 &                                                    \\
                                \cline{2-5} &     Sobel & 51.94 $\pm$ 0.82 &                                   \textbf{71.88 $\pm$ 0.76} &                  \multirow{2}{*}{\textbf{74.25 $\pm$ 0.75}} \\
                                            &   +BagNet & 42.22 $\pm$ 0.80 &                                   \textbf{73.91 $\pm$ 0.73} &                                                    \\
      \bottomrule
      \end{tabular}

        \end{center}
        \caption{CIFAR-10}
        \vspace{0.5em}
        \end{subfigure}
        \begin{subfigure}[b]{\linewidth}
        \begin{center}
            \begin{tabular}{llccc}
      \toprule
                                    Methods &  Prior(s) &     Labeled Only & \begin{tabular}{c}+Unlabeled \\Self/Co-Training\end{tabular} & \begin{tabular}{c} + Standard model \\with Pseudo-labels \end{tabular} \\
      \midrule
             \multirow{3}{*}{Self-training} &  Standard & 53.73 $\pm$ 0.94 &                                   59.92 $\pm$ 0.93 &                                   60.52 $\pm$ 0.91 \\
                                            &     Canny & 56.29 $\pm$ 0.92 &                                   58.40 $\pm$ 0.89 &                                   62.19 $\pm$ 0.91 \\
                                            &    BagNet & 52.04 $\pm$ 0.92 &                                   57.80 $\pm$ 0.99 &                                   61.69 $\pm$ 0.96 \\
      \midrule \multirow{4}{*}{Co-training} &  Standard & 53.73 $\pm$ 0.94 &                                   58.05 $\pm$ 0.95 &                  \multirow{2}{*}{61.16 $\pm$ 0.94} \\
                                            & +Standard & 55.38 $\pm$ 0.92 &                                   60.44 $\pm$ 0.92 &                                                    \\
                                \cline{2-5} &     Canny & 56.29 $\pm$ 0.94 &                                   \textbf{62.21 $\pm$ 0.93} &                  \multirow{2}{*}{\textbf{67.33 $\pm$ 0.89}} \\
                                            &   +BagNet & 52.04 $\pm$ 1.00 &                                   \textbf{66.74 $\pm$ 0.94} &                                                    \\
      \bottomrule
      \end{tabular}

        \end{center}
        \caption{STL-10}
        \end{subfigure}
    \fi

    \caption{Test accuracy of self-training and co-training methods on STL-10
    and CIFAR-10.
    For each model, we report the original accuracy when trained only
    labeled data (Column 3) as well as the accuracy after being trained on
    pseudo-labeled data (Column 4). (Recall that, for the case of
    co-training, pseudo-labeling is performed by combining the predictions of
    both models.) Finally, we report the performance of a standard model
    trained from scratch on the resulting pseudo-labels (Column 5).
    We provide 95\% confidence intervals computed via bootstrap with 5000
    iterations.
    }
    \ificml 
    \begin{subfigure}[b]{\linewidth}
        \begin{center}
            
        \end{center}
        \caption{CIFAR-10}
        \vspace{0.5em}
        \end{subfigure}
        \begin{subfigure}[b]{\linewidth}
        \begin{center}
            
        \end{center}
        \caption{STL-10}
        \end{subfigure}
        \vspace{-1.5em}
    \else
    \fi
    \label{tab:basictable}
\end{table*}

\paragraph{Co-training performance.}
We find that co-training with shape- and texture-based priors can significantly
improve test accuracy compared to self-training with any
of the priors alone (Table~\ref{tab:basictable}).
This is despite the fact that, when using self-training alone, the standard
model outperforms all other models (Column 4, Table~\ref{tab:basictable}).
Moreover, co-training models with diverse priors improves upon simply 
combining them in an ensemble (Appendix~\ref{app:ensemble_self_train}).

In Appendix~\ref{app:big_basic_table}, we report the performance of co-training
with every pair of priors. We find that co-training with shape- and texture-based priors 
(Canny + BagNet for STL-10 and Sobel + BagNet for CIFAR-10) outperforms 
every other prior combination. Note that this differs from the setting of 
ensembling models with different priors, where Standard + Sobel is consistently 
the best performing pair for CIFAR-10 (c.f Table~\ref{tab:ensembles} and
Appendix~\ref{app:ensemble_self_train}).
These results indicate that the diversity of shape- and texture-biased
models allows them to improve each other over training. 
 
Additionally, we find that, even when training a single model on the
pseudo-labels of another model, prior diversity can still help.
Specifically, we compare the performance of a standard model trained from
scratch using pseudo-labels from various self-trained models (Column 5, Table
\ref{tab:basictable}). 
In this setting, using a self-trained shape- or texture-biased model for
pseudo-labeling outperforms using a self-trained standard model.
This is despite the fact that, in isolation, the standard model has
higher accuracy than the shape- or texture-biased ones (Column 4,
Table \ref{tab:basictable}).
 
\begin{figure*}[htp]
    \hfill
    \begin{subfigure}[b]{0.39\linewidth}
        \includegraphics[width=\textwidth]{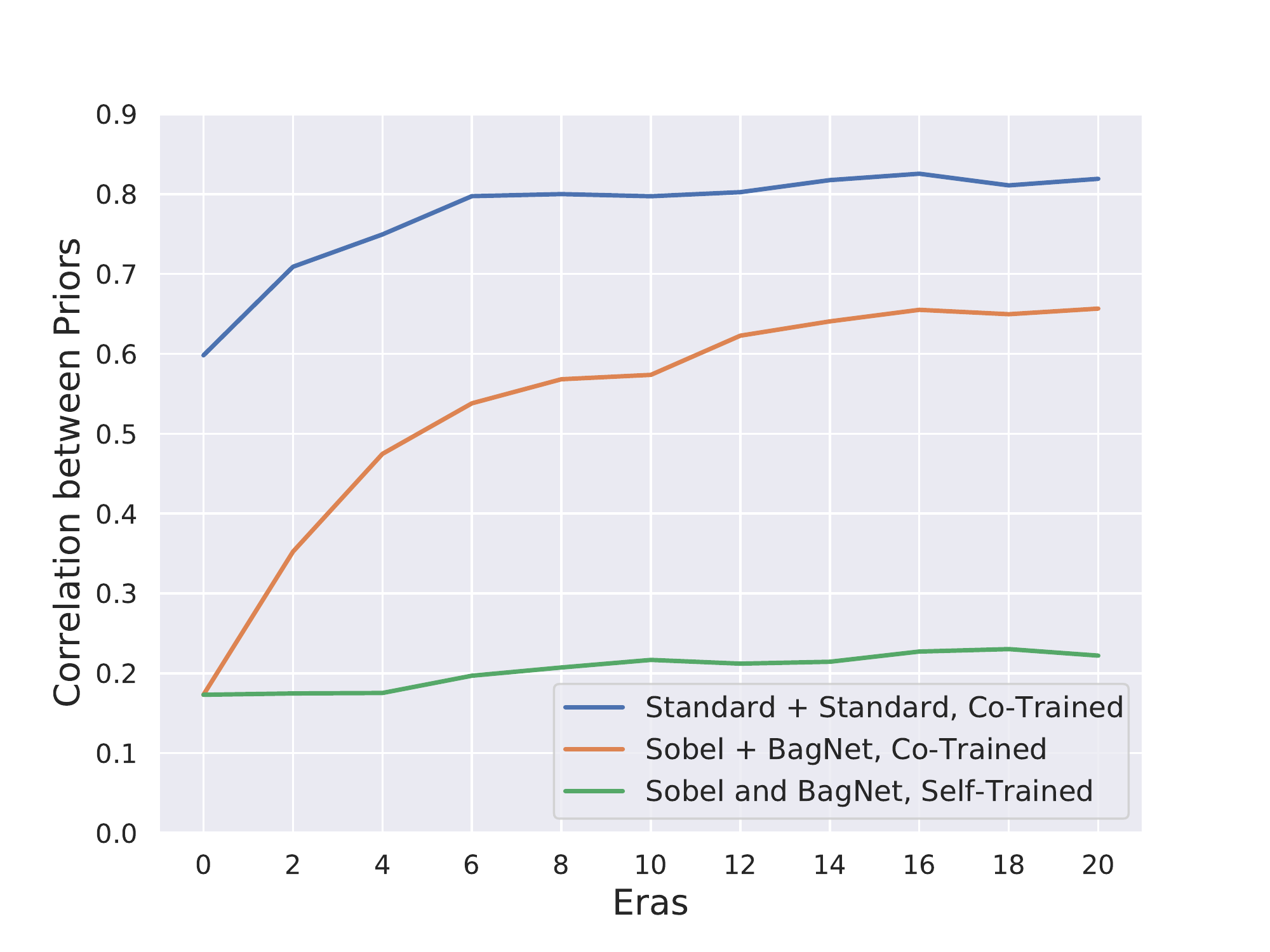}
    \caption{CIFAR-10}
    \end{subfigure}
    \hfill 
    \begin{subfigure}[b]{0.39\linewidth}
        \includegraphics[width=\textwidth]{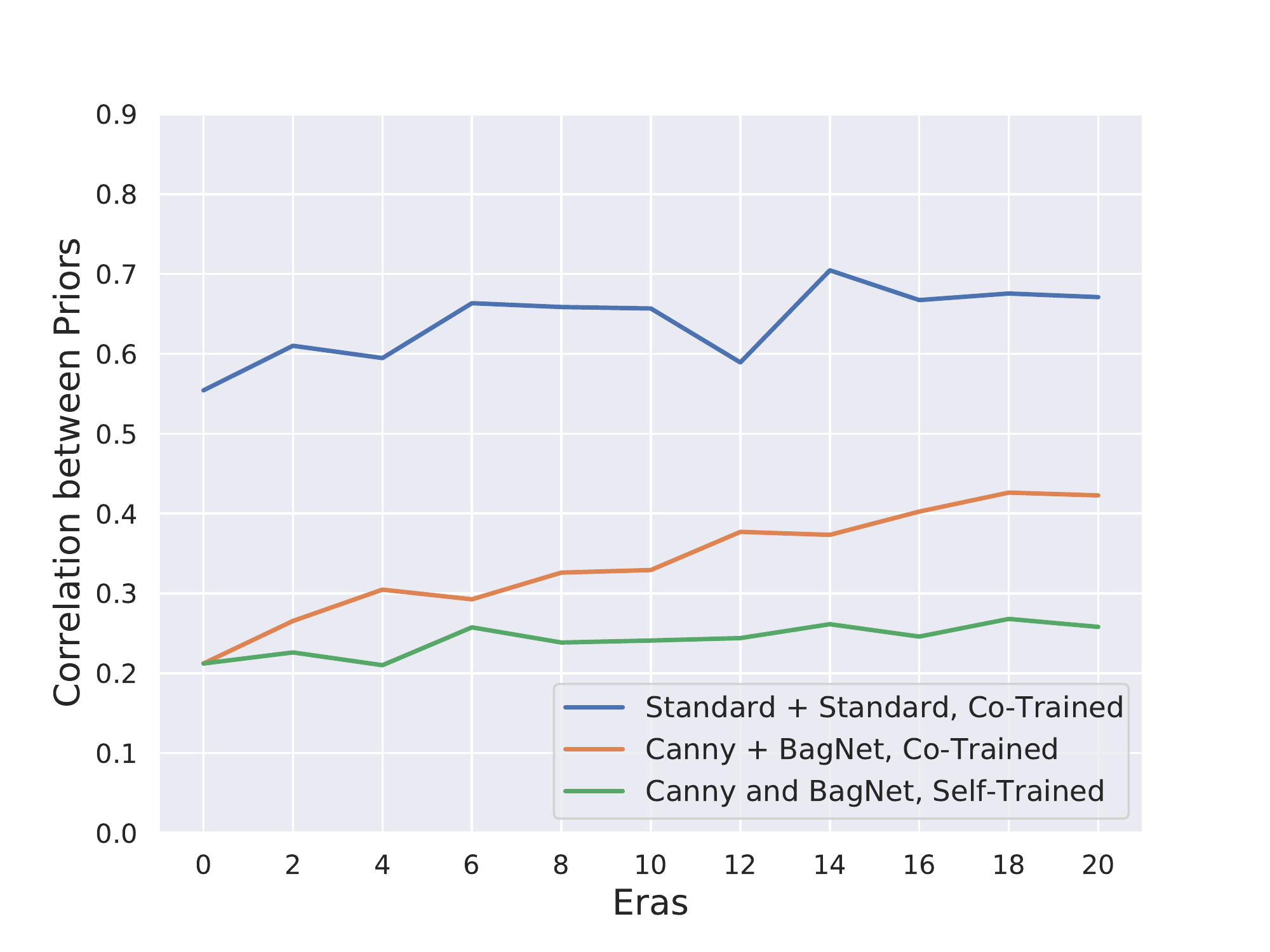}
    \caption{STL-10}
    \end{subfigure}
    \hfill 
    \phantom{}
    \caption{Correlation between the correct predictions of shape- and texture-biased models
        over the course of co-training for STL-10 and CIFAR-10. For comparison,
        we also plot the correlation between the predictions when the models
        induced by these priors are individually self-trained, as well as the
        correlation of two standard models when co-trained together.}
    \label{fig:prior_independence_over_time}
\end{figure*}

\paragraph{Model alignment over co-training.}
To further explore the dynamics of co-training, we evaluate how the correlation
between model predictions evolves as the eras progress in Figure~\ref{fig:prior_independence_over_time} 
(using the prediction alignment measure of Table~\ref{tab:independence}).
We find that shape- and texture-biased models exhibit low correlation at the
start of co-training, but this correlation increases as co-training
progresses.
This is in contrast to self-training each model on its own, where
the correlation remains relatively low.
It is also worth noting that the correlation appears to plateau at a lower value
when co-training models with distinct feature priors as opposed to co-training
two standard models.

Finally, we find that a standard model trained on the pseudo-labels of other
models correlates well with the models
themselves (see Appendix~\ref{app:independence_after_cotraining}).
Overall, these findings indicate that models trained on each other's
pseudo-labels end up behaving more similarly.

\section{Using Co-Training to Avoid Spurious Correlations}
\label{sec:spurious}
A major challenge when training models for real-world deployment is avoiding
spurious correlations: associations which are predictive on the training data
but not valid for the actual task.
Since models are typically trained to maximize train accuracy, they
are quite likely to rely on such spurious
correlations~\cite{gururangan2018annotation,beery2018recognition,geirhos2020shortcut,xiao2020noise}.

In this section, our goal is to leverage diverse feature priors to control the
sensitivity of the training process to such spurious correlations.
Specifically, we will assume that the spurious correlation does not hold on the
unlabeled data (which is likely since unlabeled data can often be collected at a larger scale).
Without this assumption, the unlabeled data contains no examples that could 
(potentially) contradict the spurious correlation (we investigate the setting where the unlabeled data 
is also similarly skewed in Appendix \ref{app:fullskew}).  
As we will see, if the problematic correlation is not easily captured by one of
the priors, the corresponding model generates pseudo-labels that are inconsistent 
with this correlation, thus steering other models away from this correlation during 
co-training. 

\paragraph{Setup.}
We study spurious correlations in two settings.
First, we create a synthetic dataset by tinting each image of the STL-10 labeled
dataset in a class-specific way. The tint is highly predictive on the training set,
but not on the test set (where this correlation is absent). 
Second, similar to \citet{sagawa2019distributionally}, we consider a gender classification task based on
CelebA~\citep{liu2015faceattributes} where hair color (``blond'' vs.\ ``non-blond'') is
predictive on the labeled data but not on the unlabeled and test data.
While gender and hair color are independent
on the unlabeled dataset, the labeled dataset consists only of blond females and non-blond
males. Similarly to the synthetic case, the labeled data encourages a prediction rule based only on
hair color.
See Appendix~\ref{app:datasets} for details.

\begin{table*}[!ht]
\ificml
\else
    \begin{subfigure}{\linewidth}
    \begin{center}
        \begin{tabular}{llccc}
    \toprule
                                  Methods & Prior(s) &     Labeled Only & \begin{tabular}{c}+Unlabeled \\Self/Co-Training\end{tabular} & \begin{tabular}{c} + Standard model \\with Pseudo-labels \end{tabular} \\
    \midrule
           \multirow{4}{*}{Self-training} & Standard & 13.99 $\pm$ 0.66 &                                   17.56 $\pm$ 0.70 &                                   17.81 $\pm$ 0.74 \\
                                          &    Canny & 55.95 $\pm$ 0.92 &                                   57.31 $\pm$ 0.89 &                                   57.81 $\pm$ 0.92 \\
                                          &    Sobel & 55.11 $\pm$ 0.91 &                                   56.12 $\pm$ 0.92 &                                   57.16 $\pm$ 0.91 \\
                                          &   BagNet & 13.10 $\pm$ 0.64 &                                   13.53 $\pm$ 0.62 &                                   14.65 $\pm$ 0.66 \\
    \midrule \multirow{4}{*}{Co-training} &    Canny & 55.95 $\pm$ 0.90 &                                   57.74 $\pm$ 0.90 &                  \multirow{2}{*}{57.85 $\pm$ 0.95} \\
                                          &  +BagNet & 13.10 $\pm$ 0.65 &
    55.33 $\pm$ 0.92 &                                                    \\
                              \cline{2-5} &    Sobel & 55.11 $\pm$ 0.95 &                                   57.71 $\pm$ 0.90 &                  \multirow{2}{*}{57.60 $\pm$ 0.94} \\
                                          &  +BagNet & 13.10 $\pm$ 0.62 &
                              54.61 $\pm$ 0.94 &                                                    \\
    \bottomrule
    \end{tabular}


    \end{center}
    \caption{Tinted STL-10}
    \vspace{0.0em}
    \end{subfigure}
    \begin{subfigure}{\linewidth}
        \begin{center}
            \begin{tabular}{llccc}
    \toprule
                                  Methods & Prior(s) &     Labeled Only & \begin{tabular}{c}+Unlabeled \\Self/Co-Training\end{tabular} & \begin{tabular}{c} + Standard model \\with Pseudo-labels \end{tabular} \\
    \midrule
           \multirow{4}{*}{Self-training} & Standard & 67.07 $\pm$ 0.58 &                                   71.57 $\pm$ 0.53 &                                   71.89 $\pm$ 0.53 \\
                                          &    Canny & 80.90 $\pm$ 0.47 &                                   85.73 $\pm$ 0.40 &                                   86.55 $\pm$ 0.42 \\
                                          &    Sobel & 82.94 $\pm$ 0.45 &                                   85.42 $\pm$ 0.43 &                                   84.96 $\pm$ 0.43 \\
                                          &   BagNet & 69.35 $\pm$ 0.55 &                                   64.89 $\pm$ 0.59 &                                   66.15 $\pm$ 0.58 \\
    \midrule \multirow{4}{*}{Co-training} &    Canny & 80.90 $\pm$ 0.46 &
    89.64 $\pm$ 0.36 &                  \multirow{2}{*}{91.99 $\pm$ 0.31} \\
                                          &  +BagNet & 69.35 $\pm$ 0.55 &
            91.44 $\pm$ 0.33 &                                                    \\
                              \cline{2-5} &    Sobel & 82.94 $\pm$ 0.44 &
                              90.64 $\pm$ 0.35 &                  \multirow{2}{*}{
                              90.99 $\pm$ 0.34} \\
                                          &  +BagNet & 69.35 $\pm$ 0.57 &
                              88.72 $\pm$ 0.39 &                                                    \\
    \bottomrule
    \end{tabular}

        \end{center}
    \vspace{-0.3em}
        \caption{CelebA}
    \end{subfigure}
\fi
\caption{Test accuracy of self-training and co-training on tinted STL-10 and CelebA,
    two datasets with spurious features (table structure is identical
    Table~\ref{tab:basictable}).
In both datasets, the spurious correlation is
more easily captured by the BagNet and Standard models over the shape-based ones.
Nevertheless, when co-trained with a shaped-biased model, BagNets are able to
significantly improve their performance, indicating that they rely less on this
spurious correlation. CI: 95\% bootstrap.}
\ificml
    \begin{subfigure}{\linewidth}
    \begin{center}
        
    \end{center}
    \caption{Tinted STL-10}
    \vspace{0.0em}
    \end{subfigure}
    \begin{subfigure}{\linewidth}
        \begin{center}
            
        \end{center}
        \caption{CelebA}
    \end{subfigure}
    \vspace{-1em}
\else
\fi
\label{tab:spurioussynthetic}
\end{table*}

\paragraph{Performance on datasets with spurious features.}
We find that, when trained only on the labeled data (where the correlation is
fully predictive), both the standard and BagNet models generalize poorly in
comparison to the shape-biased models (see Table~\ref{tab:spurioussynthetic}).
This behavior is expected: the spurious attribute in both datasets is
color-related and mostly suppressed by the edge detection algorithms used
to train shape-based models.
Even after self-training on the unlabeled data (where the correlation
is absent), the performance of the standard and BagNet models does not improve
significantly.
Finally, simply ensembling self-trained models post hoc does not improve their
performance.
Indeed as the texture-biased and standard models are significantly less accurate
than the shape-biased one, they lower the overall accuracy of the
ensemble (see Appendix~\ref{app:spurious_ensembles}).

In contrast, when we co-train shape- and texture- biased models together,
the texture-biased model improves substantially.
When co-trained with a Canny model, the BagNet model
improves over self-training by 42\% on the tinted
STL-10 dataset and 27\% on the CelebA dataset. 
This improvement can be attributed to the fact that the predictions of the
shape-biased model are inconsistent with the spurious correlation on the
unlabeled data.  
By being trained on pseudo-labels from that model, the BagNet model is
forced to rely on alternative, non-spurious features.

Moreover, particularly on CelebA, the shape-biased model also
improves when co-trained with a texture-biased model.
This indicates that even though the texture-biased model relies on the
spurious correlation, it also captures non-spurious features that, during co-training, improve the performance of the shape-based model.
In Appendix~\ref{app:splitceleba}, we find that these improvements 
are concentrated on inputs
where the spurious correlation does not hold.

\section{Additional Related Work}
\label{sec:related}
In Section \ref{sect:priors}, we discussed the most relevant prior work on
implicit or explicit feature priors. Here, we discuss additional related work
and how it connects to our approach.

\paragraph{Shape-biased models.}
Several other methods aim to bias models towards shape-based
features:
input stylization~\cite{geirhos2018imagenettrained,somavarapu2020frustratingly,li2021shapetexture},
penalizing early layer predictiveness~\cite{wang2019learning},
jigsaw puzzles~\cite{carlucci2019domain, asadi2019towards},
dropout~\cite{shi2020informative},
or data augmentation~\cite{hermann2020origins}.
While, in our work, we choose to suppress texture information via edge detection
algorithms, any of these methods can be
substituted to generate the shape-based model for our analysis.

\paragraph{Avoiding spurious correlations.}
Other methods to avoid learning spurious correlations
include:
 learning representations that are
 optimal across domains~\citep{arjovsky2019invariant},
enforcing robustness to group shifts~\citep{sagawa2019distributionally}, and
utilizing multiple data points corresponding to each physical
entity~\citep{heinze2017conditional}.
Similar in spirit to our work, these methods encourage prediction rules that are
supported by multiple views of the data.
However, we do not rely on annotations or multiple sources and
instead impose feature priors through the model architecture and
input preprocessing.

\paragraph{Pseudo-labeling.} 
Since the initial proposal of pseudo-labeling for neural networks~\cite{lee2013pseudo},
there has been a number of more sophisticated pseudo-labeling schemes 
aimed at improving the accuracy and diversity of the labels~\cite{iscen2019label,augustin2020out,xie2020self,rizve2021defense,huang2021self}.
In our work, we focus on the simplest scheme for self-labeling---i.e.,
confidence based example selection.
Nevertheless, most of these schemes can be directly incorporated into our
framework to potentially improve its overall performance.

A recent line of work explores  self-training
by analyzing it under different assumptions on the
data~\citep{mobahi2020self,wei2020theoretical,allen2020towards,kumar2020understanding}.
Closest to our work, \citet{chen2020self} show that self-training on unlabeled
data can reduce reliance on spurious correlations under certain assumptions.
In contrast, we demonstrate that by leveraging diverse feature priors, we can
avoid spurious correlations even if a model heavily relies on them.

\paragraph{Consistency regularization.}
Consistency regularization, where a model is 
trained to be invariant to a set of input transformations, is another canonical
technique for leveraging unlabeled data. 
These transformations might stem from data augmentations and architecture
stochasticity~\cite{laine2017temporal,berthelot2019mixmatch,chen2020simple,sohn2020fixmatch,prabhu2021sentry}
or using adversarial examples~\cite{miyato2018virtual}.

\paragraph{Ensemble diversity.}
While the standard recipe for creating model ensembles is based on training
multiple identical models from different random
initializations~\citep{lakshminarayanan2017simple}, there do exist other
methods for introducing diversity.
Examples include training models with different
hyperparameters~\citep{wenzel2020hyperparameter}, data
augmentations~\citep{stickland2020diverse}, input
transformations~\citep{yeo2021robustness}, or model
architectures~\citep{zaidi2020neural}.
Note that, in contrast to our work, none of these approaches incorporate
this diversity into training itself.

\paragraph{Co-training.}
One line of work studies co-training from a
theoretical 
perspective~\citep{nigam2000analyzing,balcan2005co,goldman2000enhancing}.
Other work aims to improve co-training by either expanding the settings where it can be applied~\citep{chen2011automatic} or by improving its
stability~\citep{ma2020self,zhang2011cotrade}.
Finally, a third line of work applies co-training to images.
Since images cannot be separated into disjoint feature sets,
one applies co-training by training multiple
models~\cite{han2018co}, either regularized to be
diverse through adversarial examples~\cite{qiao2018deep} or trained using different
methods~\cite{yang2020mico}.
Our method is complementary to these approaches as it relies on 
explicit feature priors to obtain different views.

\section{Conclusion}
In this work, we explored the benefits of combining feature priors with non-overlapping
failure modes. By capturing complementary perspectives on the data, 
models trained with diverse feature priors can offset each other's mistakes when 
combined through methods such as ensembles. Moreover, in the presence of unlabeled 
data, we can leverage prior diversity by jointly boostrapping models with different 
priors through co-training. This allows the models to correct each other during 
training, thus improving pseudo-labeling and controlling for correlations that 
do not generalize well.

We believe that our work is only the first step in exploring the design space
of creating, manipulating, and combining feature priors to 
improve generalization.  In particular, our framework is quite flexible and allows 
for a number of different design choices, such as choosing other feature priors (cf.
Sections~\ref{sec:priors} and~\ref{sec:related}), using other methods for 
pseudo-label selection (e.g., using uncertainty estimation \citep{lee2018simple,rizve2021defense}),
and combining pseudo-labels via different ensembling
methods. More broadly, we believe that exploring the synthesis of explicit feature priors 
in new applications is an exciting avenue for further research.

\section*{Acknowledgements}
We thank Shibani Santurkar for helpful discussions.

Work supported in part by the NSF grants CCF-1553428 and CNS-1815221, Open Philanthropy, and a Facebook Fellowship. This material is based upon work supported by the Defense Advanced Research Projects Agency (DARPA) under Contract No. HR001120C0015.

Research was sponsored by the United States Air Force Research Laboratory and the United States Air Force Artificial Intelligence Accelerator and was accomplished under Cooperative Agreement Number FA8750-19-2-1000. The views and conclusions contained in this document are those of the authors and should not be interpreted as representing the official policies, either expressed or implied, of the United States Air Force or the U.S. Government. The U.S. Government is authorized to reproduce and distribute reprints for Government purposes notwithstanding any copyright notation herein.

\printbibliography
\clearpage

\appendix

\section{Experimental Details} 
\label{app:setup}
\subsection{Datasets}
\label{app:datasets}

For our first set of experiments (Section~\ref{sect:methodology}), we focus on a
canonical setting where a small portion of the training set if labeled and we
have access to a pool of unlabeled data.

\paragraph{STL-10.}
The STL-10~\cite{coates2011analysis} dataset contains 5,000 training
and 8,000 test images of size 96$\times$96 from 10 classes.
We designate 1,000 of the 5,000 (20\%) training examples to be the labeled
training set, 500 (10\%) to be the validation set, and the rest are used as
unlabeled data.

\paragraph{CIFAR-10.}
The CIFAR-10~\cite{krizhevsky2009learning} dataset contains 50,000
training and 8,000 test images of size 32$\times$32 from 10 classes.
We designate 1,000 of the 50,000 (2\%) training examples to be the labeled
training set, 5000 (10\%) to be the validation set, and the rest as unlabeled
data.\\

In both cases, we report the final performance on the standard test set of
that dataset.
We also create two datasets that each contain a different spurious correlation.

\begin{figure}[htp]
    \begin{subfigure}[b]{\linewidth}
        \centering
        \includegraphics[width=.95\linewidth]{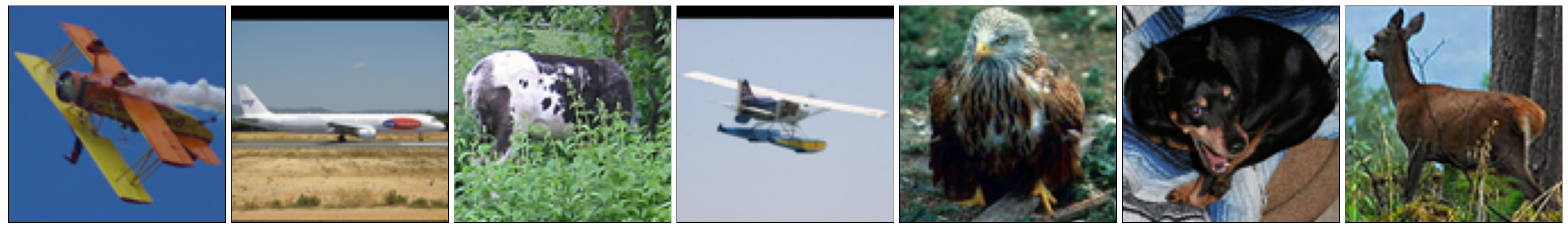}
        \caption{Original}
    \end{subfigure}
    \begin{subfigure}[b]{\linewidth}
        \centering
        \includegraphics[width=.95\linewidth]{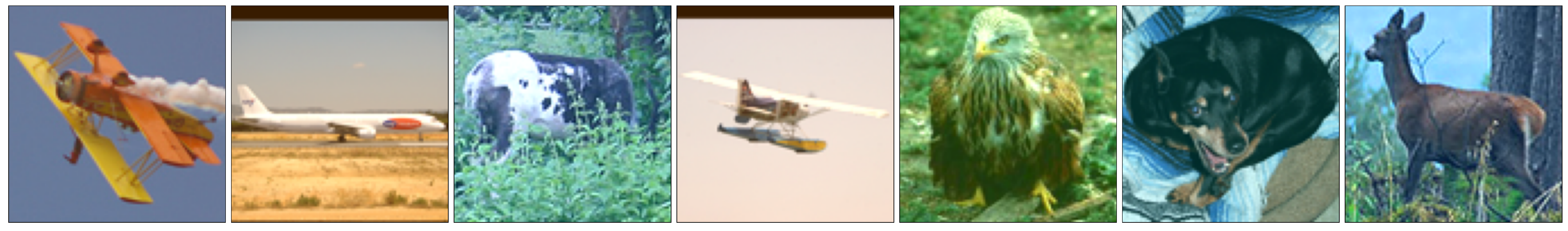}
        \caption{Tinted}
    \end{subfigure}
    \caption{Tinted STL-10 images. The tint is
    class-specific and thus models can learn to predict based mostly on that
    tint.}
    \label{fig:tint}
\end{figure}

\paragraph{Tinted STL-10.}
We reuse the STL-10 setup described above, but we add a class-specific tint to
each image in the (labeled) training set.
Specifically, we hand-pick a different color for each of the 10 classes and then
add this color to each of the pixels (ensuring that each RGB channel remains
within the valid range)---see Figure~\ref{fig:tint} for examples.
This tint is only present in the labeled part of the training set, the unlabeled
and test parts of the dataset are left unaltered.

\paragraph{Biased CelebA.}
We consider the task of predicting gender in the
CelebA~\cite{liu2015faceattributes} dataset.
In order to create a biased training set, we choose a random sample of 500
non-blond males and 500 blond females.
We then use a balanced unlabeled dataset consisting of 1,000 random samples for
each of: blond males, blond females, non-blond males, and non-blond females.
We use the standard CelebA test set which consists of 12.41\% blond females,
48.92\% non-blond females, 0.90\% blond males, and 37.77\% non-blond males.
(Note that a classifier predicting purely based on hair color with have an
accuracy of 50.18\% on that test set.)\\

All of the datasets that we use are freely available for non-commercial
research purposes.
Moreover, to the best of our knowledge, they do not contain offensive content or
identifiable information (other than publicly available celebrity photos).

\subsection{Model architectures and input preprocessing}
\label{app:models}

\begin{figure}[htp]
    \begin{center}
    \hfill
    \begin{subfigure}[b]{0.15\linewidth}
        \includegraphics[width=\linewidth]{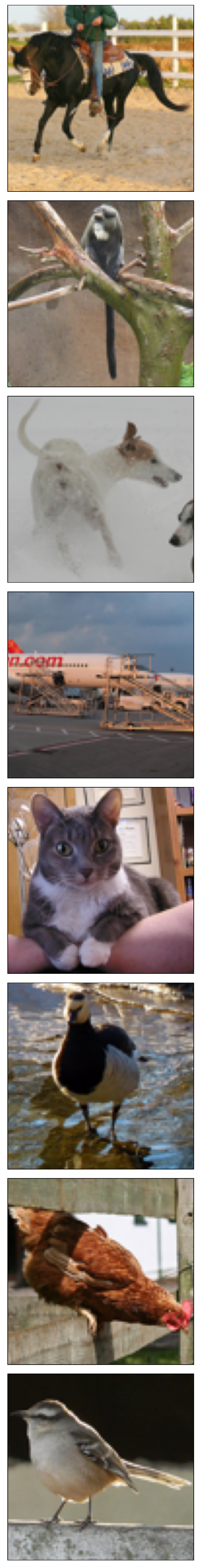}
        \caption{Original}
    \end{subfigure}\hfill
    \begin{subfigure}[b]{0.15\linewidth}
        \includegraphics[width=\linewidth]{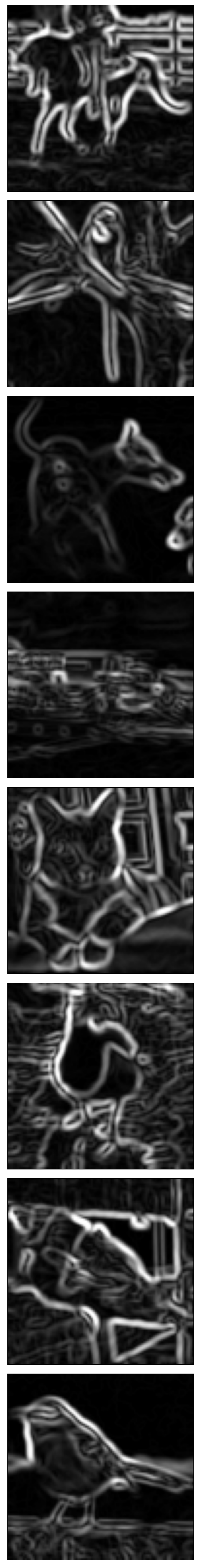}
        \caption{Sobel}
    \end{subfigure}\hfill
    \begin{subfigure}[b]{0.15\linewidth}
        \includegraphics[width=\linewidth]{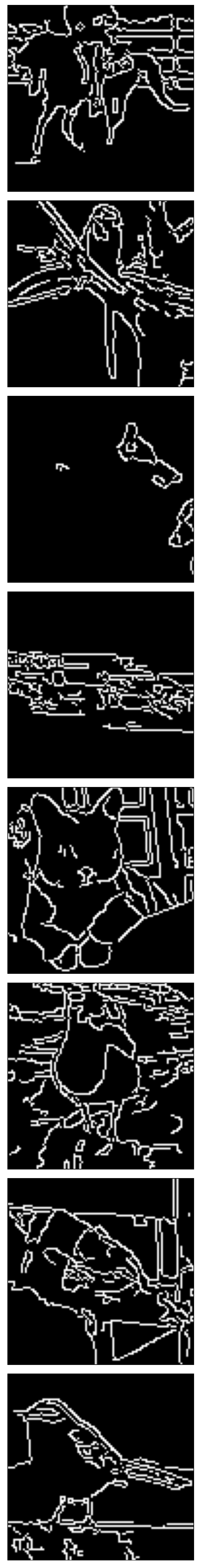}
        \caption{Canny}
    \end{subfigure}\hfill
    \begin{subfigure}[b]{0.15\linewidth}
        \includegraphics[width=\linewidth]{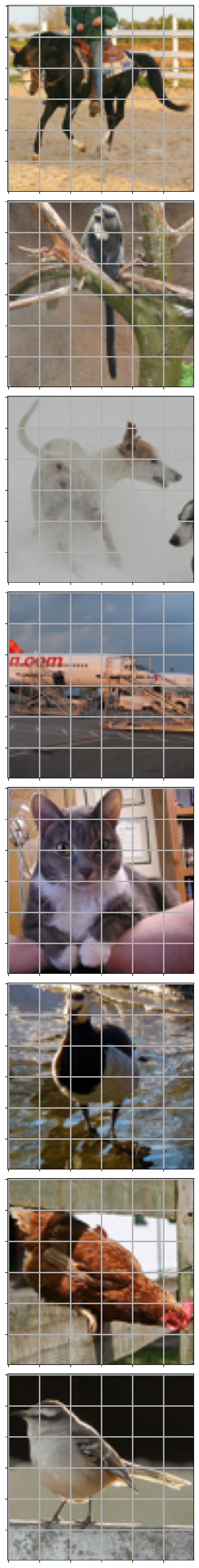}
        \caption{BagNet}
    \end{subfigure}\hfill\phantom{}
    \end{center}
    \caption{Further visualizations of the different feature priors we
    introduce. For each original image (a), we visualize the output of both edge
    detection algorithms---Sobel (b) and Canny (c)---as well as the receptive
    field of the BagNet model.}
    \label{fig:moreexamples}
\end{figure}

For both the standard model and the models trained on images processed by edge
detection algorithm, we use a standard model architecture---namely,
VGG16~\cite{simonyan2015very} with the addition of batch
normalization~\cite{ioffe2015batch} (often referred to as VGG16-BN).
We describe the exact edge detection process as well as the architecture of the
BagNet model (texture prior) below.
We visualize these priors in Figure~\ref{fig:moreexamples}.

\paragraph{Canny edge detection.}
Given an image, we first smooth it with a 5 pixel bilateral filter
\cite{tomasi1998bilateral}, with filter $\sigma$ in the coordinate and color
space set to 75.
After smoothing, the image is converted to gray-scale.
Finally, a Canny filter \cite{canny1986computational} is applied to the
image, with hysteresis thresholds 100 and 200, to extract the edges.

\paragraph{Sobel edge detection.}
Given an image, we first upsample it to 128$\times$128 pixels.
Then we convert it to gray-scale and apply a Gaussian blur (kernel size=5,
$\sigma=5$).
The image is then passed through a Sobel filter \cite{sobel19683x3} with a
kernel size of 3 in both the horizontal and the vertical direction to extract
the image gradients.

\begin{figure}[!htp]
    \centering
    \includegraphics[width=0.5\textwidth]{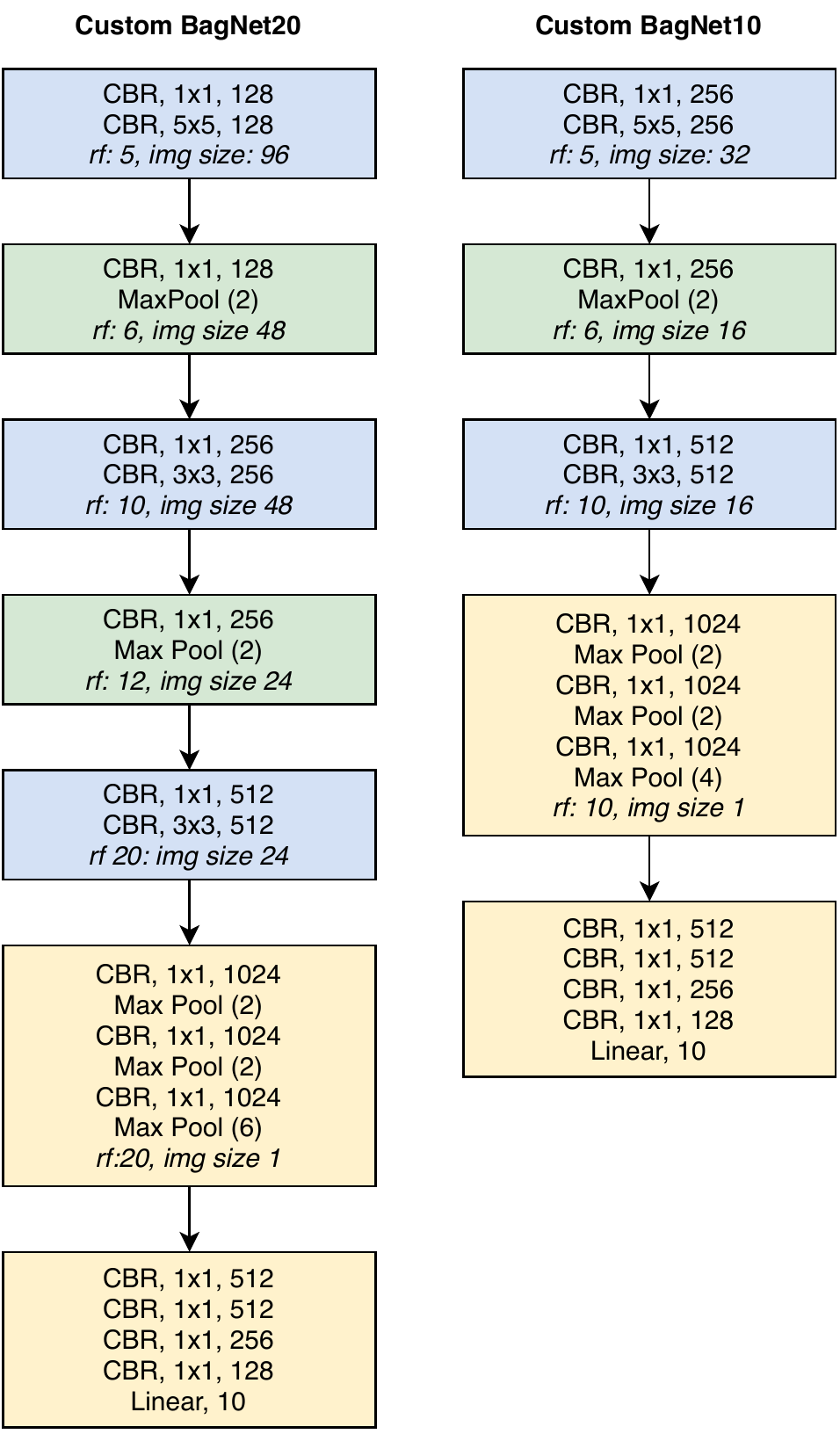}
    \caption{The customized BagNet architecture used for training texture-biased
        models.
        The basic building block consists of a convolutional layer, followed by
        batch normalization and finally a ReLU non-linearity (denoted
    collectively as \emph{CBR}).}
    \label{fig:bagnet_arch}
\end{figure}

\paragraph{BagNet.}
For our texture-biased model, we use a slimmed down version of the BagNet
architecture from \citet{brendel2018approximating}.
The goal of this architecture is to limit the receptive field of the model,
hence forcing it to make predictions based on local features.
The exact architecture we used is shown in Figure~\ref{fig:bagnet_arch}.
Intuitively, the top half of the network---i.e., the green and blue
blocks---construct features on patches of size 20$\times$20 for 96$\times$96
images and 10$\times$10 for 32$\times$32 images.
The rest of the network consists only of 1$\times$1 convolutions and
max-pooling, hence not utilizing the image's spatial structure.

\clearpage  
\subsection{Training setup}
\label{app:training}

\subsubsection{Basic training}
\label{app:basic}
We train all our models using stochastic gradient descent (SGD) with momentum
(a coefficient of 0.9) and a decaying learning rate.
We add weight decay regularization with a coefficient of $10^{-4}$.
In terms of data augmentation, we apply random cropping with a padding of 4
pixels, random horizontal flips, and a random rotation of $\pm 2$ degrees.
These transformations are applied after the edge detection processing.
We train all models with a batch size of 64 for 96$\times$96-sized images and
128 for 32$\times$32-sized images for a total of 300 epochs.
All our experiments are performed using our internal cluster which mainly
consists of NVIDIA 1080 Ti GTX GPUs.

\paragraph{Hyperparameter tuning.}
To ensure a fair comparison across feature priors, we selected the
hyperparameters for each dataset-prior pair separately, using the held-out
validation set (separate from the final test used for reporting performance).
Specifically, we performed a grid search choosing the learning rate (LR) from
$[0.1, 0.05, 0.02, 0.01, 0.005]$, the number of epochs between each learning
rate drop ($K$) from $[50, 100, 300]$ and the factor with which the learning
rate is multiplied ($\gamma$) from $[0.5, 1]$.
The parameters chosen are shown in Table~\ref{tab:hyperparams}.
We found that all models achieved near-optimal performance strictly within the
range of each hyperparameters. Thus, we did not consider a wider grid.

\begin{table}[h]
    \begin{center}
        \begin{tabular}{llccc}
    \toprule
    Dataset & Prior & LR & $\gamma$ & $K$\\
    \midrule
    \multirow{4}{*}{STL-10}& Standard & 0.01 &   0.5 & 100\\
                           & Canny    & 0.01 &   0.5 & 100\\
                           & Sobel    & 0.005&   0.5 & 100\\
                           & BagNet   & 0.05 &   0.5 & 100\\ \midrule
    \multirow{4}{*}{CIFAR-10} & Standard & 0.01 &   0.5 & 100\\
                           & Canny    & 0.01 &   0.5 & 100\\
                           & Sobel    & 0.01 &   0.5 & 100\\
                           & BagNet   & 0.01 &    0.1 & 100\\ \midrule
    \multirow{4}{*}{CelebA}& Standard & 0.005 &      0.5 & 50\\ 
                           & Canny    & 0.005 &     0.1  & 100\\
                           & Sobel    & 0.01  &     0.5  & 50\\
                           & BagNet   & 0.02  &     0.5  & 100\\
    
    \bottomrule
\end{tabular}

    \end{center}
    \caption{Hyperparameters chosen through grid search for each dataset-prior
        pair (we used the STL-10 hyperparameters for the tinted STL-10 dataset).
        LR corresponds to the learning rate, $\gamma$ to the factor used to
        decay the learning rate at each drop, and $K$ to the train epochs
        between each learning rate drop.  }
    \label{tab:hyperparams}
\end{table}

\subsection{Ensembles}
\label{app:ensemble}
In order to leverage prior diversity, we ensemble models trained with (potentially)
different priors. We use the following ensembles:
\begin{enumerate}
    \item \textbf{Take Max:} Predict based on the model assigning the highest
        probability on this example.
    \item \textbf{Average:} Average the (softmax) output probabilities of the
        models, predict the class assigned the highest probability.
    \item \textbf{Rank:} Each model ranks all test examples based on the
        probability assigned to their predicted labels.
        Then, for each example, we predict using the model which has a lower
        rank on this example.
\end{enumerate}

We then report the maximum of these ensemble methods in Table~\ref{tab:ensembles}. We separately
examine a more complex ensembling method (stacked ensembling) in Appendix~\ref{app:stacked_ensembling}.

\subsection{Self-training and co-training schemes}
\label{app:co_training_algos}
In the setting that we are focusing on, we are provided with a
labeled dataset $\mathbf{X}$ and an unlabeled dataset $\mathbf{U}$, where
typically there is much more unlabeled data ($|\mathbf{U}|\gg|\mathbf{X}|$).
We are then choosing a set of (one or more) feature priors each of which
corresponds to a different way of training a model (e.g., using edge detection
preprocessing).

\paragraph{General methodology.}
We start by training each of these models on the labeled dataset.
Then, we combine the predictions of these models to produce pseudo-labels for
the unlabeled dataset.
Finally, we choose a fraction of the unlabeled data and train the models on that
set using the produced pseudo-labels (in additional to the original labeled set
$X$).
This process is repeated using increasing fractions of the unlabeled dataset
until, eventually, models are trained on its entirety.
We refer to each such phase as an \emph{era}.
We include an additional 5\% of the unlabeled data per era, resulting
in a total of 20 eras.
During each era, we use the training process described in
Appendix~\ref{app:basic} without re-initializing the models (warm start).
After completing this process, we train a standard model from scratch using both
the labeled set and resulting pseudo-labels.
The methodology used for choosing and combining pseudo-labels is described below
for each scheme.

\paragraph{Self-training.}
Since we are only training one model, we only need to decide how to choose the
pseudo-labels to use for each era.
We do this in the simplest way: at ear $t$, we pick the subset $\mathbf{U_t}
\subseteq \mathbf{U}$ of examples that are assigned the highest probability on
their predicted label.
We attempt to produce a class-balanced training set by applying this process
separately on each class (as predicted by the model).
The pseudocode for the method is provided in Algorithm~\ref{algo:selftraining}.

\ificml
\begin{algorithm}[!htbp]
    \caption{Self-Training}
    \label{algo:selftraining}
 \begin{algorithmic}
    \STATE {\bfseries Params:} Number of eras $T$. Fraction added per era $k$.
    \STATE {\bfseries Input:} Labeled data $\mathbf{X}$ with $n$ classes, unlabeled data
    $\mathbf{U}$, model trained on $\mathbf{X}$.
    \FOR{era\ $t \in 1...T$}
        \STATE forward-pass $\mathbf{U}$ through model to create pseudo-labels;
        \STATE $\mathbf{U_t} = []$;
        \FOR{each class $c$}
            \STATE Select the $\frac{kt|\mathbf{U}|}{n}$ most confident examples from $\mathbf{U}$ predicted by the model as class $c$;
            \STATE Add those examples to $\mathbf{U_t}$ with class $c$;
        \ENDFOR
        \STATE Re-train (warm start) the model on $\mathbf{X} \cup \mathbf{U_t}$ until convergence;
    \ENDFOR
    \STATE Train a standard model from scratch on $\mathbf{X} \cup \mathbf{U_T}$.
 \end{algorithmic}
 \end{algorithm}
\else
\begin{algorithm}[htp]
    \DontPrintSemicolon
    \Parameters{Number of eras $T$. Fraction added per era $k$.}
    \Input{Labeled data $\mathbf{X}$ with $n$ classes, unlabeled data
    $\mathbf{U}$, model trained on $\mathbf{X}$.}
     \For{era\ $t \in 1...T$}{
         forward-pass $\mathbf{U}$ through the model to create pseudo-labels\;
         $\mathbf{U_t} = []$\;
         \For{each class $c$\ }{
            Select the $\frac{kt|\mathbf{U}|}{n}$ most confident examples from $\mathbf{U}$ predicted by the model as class $c$\;
            Add those examples to $\mathbf{U_t}$ with class $c$\;
         }
         Re-train (warm start) the model on $\mathbf{X} \cup \mathbf{U_t}$ until convergence\;
     }
     Train a standard model from scratch on $\mathbf{X} \cup \mathbf{U_T}$.
     \caption{Self-training}
     \label{algo:selftraining}
\end{algorithm}
\fi

\clearpage
\paragraph{Standard co-training.}
Here, we train multiple models (in our experiments two) based on a common pool
of pseudo-labeled examples in each era.
In each era $t$, each model labels the unlabeled dataset $\mathbf{U}$.
Then, for each class, we alternate between models, adding the next most
confident example predicted as that class for that model to $\mathbf{U_t}$,
until we reach a fixed number of unique examples have been added for that class
(5\% of the size of the unlabeled dataset per era).
Note that this process allows both conflicts and duplicates: if multiple models
are confident about a specific example, that example may be added more than once
(potentially with a different label each time).
Finally, we train each model (without re-initializing) on $\mathbf{X} \cup
\mathbf{U_t}$.
The pseudocode for this method can be found in Algorithm~\ref{algo:stdcotrain}.

\ificml
\begin{algorithm}[htp]
    \caption{Standard Co-Training}
    \label{algo:stdcotrain}
 \begin{algorithmic}
    \STATE {\bfseries Params:} Number of eras $T$. Fraction added per era $k$.
    \STATE {\bfseries Input:} Labeled data $\mathbf{X}$ with $n$ classes, unlabeled data
    $\mathbf{U}$, models trained on $\mathbf{X}$.
     \FOR{era $t \in 1...T$}
         \STATE forward-pass $\mathbf{U}$ through each model to create pseudo-labels;
         \STATE $\mathbf{U_t} = []$;
         \FOR{each class $c$}
             \STATE $\mathbf{U_t^{(c)}} = []$;
             \WHILE{the \# of unique examples in $\mathbf{U_t^{(c)}} < \frac{kt|\mathbf{U}|}{n}$}
                \FOR{each model $m$} 
                    \STATE Add the next most confident example predicted by $m$ as class $c$ to $\mathbf{U_t^{(c)}}$;
                \ENDFOR 
             \ENDWHILE
            \STATE Add $\mathbf{U_t^{(c)}}$ to $\mathbf{U_t}$;
         \ENDFOR
         \STATE Re-train (warm start) each model on $\mathbf{X} \cup \mathbf{U_t}$
         until convergence;
     \ENDFOR
     \STATE Train a standard model from scratch on $\mathbf{X} \cup \mathbf{U_T}$.
 \end{algorithmic}
 \end{algorithm}
\else
\begin{algorithm}[ht]
    \DontPrintSemicolon
    \Parameters{Number of eras $T$. Fraction added per era $k$.}
    \Input{Labeled data $\mathbf{X}$ with $n$ classes, unlabeled data
    $\mathbf{U}$, models trained on $\mathbf{X}$.}
     \For{era $t \in 1...T$}{
         forward-pass $\mathbf{U}$ through each model to create pseudo-labels\;
         $\mathbf{U_t} = []$\;
         \For{each class $c$}{
             $\mathbf{U_t^{(c)}} = []$\;
             \While{the number of unique examples in $\mathbf{U_t^{(c)}} < \frac{kt|\mathbf{U}|}{n}$} {
                \For{each model $m$} {
                    Add the next most confident example predicted by $m$ as
                    class $c$ to $\mathbf{U_t^{(c)}}$\;
                 }
             }
            Add $\mathbf{U_t^{(c)}}$ to $\mathbf{U_t}$
         }
         Re-train (warm start) each model on $\mathbf{X} \cup \mathbf{U_t}$
         until convergence\;
     }
     Train a standard model from scratch on $\mathbf{X} \cup \mathbf{U_T}$.
     \caption{Standard Co-Training}
     \label{algo:stdcotrain}
\end{algorithm}
\fi

\clearpage
\section{Additional Experiments} 
\subsection{Experiment organization}
We now provide the full experimental results used to create the plots in the
main body as well as additional analysis.
Specifically, in Appendix~\ref{app:full_pretrained_ensemble}
and~\ref{app:ensemble_self_train} we present the performance of individual
ensemble schemes for pre-trained and self-trained models respectively.
Then, in Appendix~\ref{app:big_basic_table} we present the performance of
co-training for each combination of feature priors.
In Appendix~\ref{app:independence_after_cotraining} we analyse the effect that
co-training has on model similarity after training.
Finally, in Appendix~\ref{app:spurious_ensembles} we evaluate model ensembles on
datasets with spurious correlations and in Appendix~\ref{app:splitceleba}
we breakdown the performance of co-training on the skewed CelebA dataset
according to different input attributes.

\subsection{Full Pre-Trained Ensemble Results}
\label{app:full_pretrained_ensemble}
In Table~\ref{tab:ensembles}, we reported the best ensemble method for each pair of models 
trained with different priors on the labeled data.  In Table~\ref{tab:big_ensemble_baselines},
we report the full results over the individual ensembles. 
\begin{table}[!h]
    \centering
    \begin{subfigure}[b]{\textwidth}
        \centering
        \begin{footnotesize}
  \addtolength{\tabcolsep}{-1pt}    
  \begin{tabular}{lcccccc}
    \toprule
         Feature Priors &          Model 1 &          Model 2 &     Max Conf. &    Avg Conf. &        Rank &        Best \\
    \midrule
    Standard + Standard & 52.54 $\pm$ 0.85 & 51.82 $\pm$ 0.85 & 53.98 $\pm$ 0.83 & 54.02 $\pm$ 0.85 & 53.98 $\pm$ 0.83 & 54.02 $\pm$ 0.82 \\
          Sobel + Sobel & 51.94 $\pm$ 0.88 & 53.69 $\pm$ 0.86 & 54.62 $\pm$ 0.83 & 54.68 $\pm$ 0.86 & 54.61 $\pm$ 0.85 & 54.68 $\pm$ 0.83 \\
          Canny + Canny & 45.48 $\pm$ 0.84 & 44.19 $\pm$ 0.88 & 46.46 $\pm$ 0.82 & 46.48 $\pm$ 0.86 & 46.70 $\pm$ 0.83 & 46.70 $\pm$ 0.79 \\
        BagNet + BagNet & 42.22 $\pm$ 0.80 & 42.56 $\pm$ 0.83 & 43.32 $\pm$ 0.82 & 43.49 $\pm$ 0.82 & 43.33 $\pm$ 0.85 & 43.49 $\pm$ 0.84 \\
        \hline
       Standard + Sobel & 52.54 $\pm$ 0.79 & 51.94 $\pm$ 0.82 & 58.14 $\pm$ 0.82 & 58.21 $\pm$ 0.88 & 58.12 $\pm$ 0.82 & \textbf{58.21 $\pm$ 0.90} \\
       Standard + Canny & 52.54 $\pm$ 0.87 & 45.48 $\pm$ 0.81 & 55.18 $\pm$ 0.82 & 55.49 $\pm$ 0.83 & 54.41 $\pm$ 0.81 & 55.49 $\pm$ 0.83 \\
      Standard + BagNet & 52.54 $\pm$ 0.85 & 42.22 $\pm$ 0.80 & 52.89 $\pm$ 0.84 & 53.03 $\pm$ 0.89 & 50.69 $\pm$ 0.81 & 53.03 $\pm$ 0.85 \\
          Sobel + Canny & 51.94 $\pm$ 0.82 & 45.48 $\pm$ 0.85 & 53.81 $\pm$ 0.84 & 53.95 $\pm$ 0.80 & 53.18 $\pm$ 0.91 & 53.95 $\pm$ 0.85 \\
         Sobel + BagNet & 51.94 $\pm$ 0.86 & 42.22 $\pm$ 0.82 & 54.42 $\pm$ 0.84 & 55.14 $\pm$ 0.83 & 53.50 $\pm$ 0.82 & 55.14 $\pm$ 0.84 \\
         Canny + BagNet & 45.48 $\pm$ 0.78 & 42.22 $\pm$ 0.79 & 49.95 $\pm$ 0.84 & 50.57 $\pm$ 0.82 & 49.64 $\pm$ 0.81 & 50.57 $\pm$ 0.84 \\
    \bottomrule
    \end{tabular}
    
    \addtolength{\tabcolsep}{1pt}    
\end{footnotesize}

    \caption{Ensemble Baselines for CIFAR-10}
    \end{subfigure}

    \begin{subfigure}[b]{\textwidth}
        \centering
        \begin{footnotesize}
\addtolength{\tabcolsep}{-1pt}    
\begin{tabular}{lccccccc}
  \toprule
       Feature Priors &          Model 1 &          Model 2 &   Max Conf. &    Avg Conf. &        Rank &        Best \\
  \midrule
  Standard + Standard & 53.73 $\pm$ 0.86 & 55.38 $\pm$ 1.00 & 56.95 $\pm$ 0.94 & 57.06 $\pm$ 0.91 & 56.94 $\pm$ 0.97 & 57.06 $\pm$ 0.91 \\
        Sobel + Sobel & 55.49 $\pm$ 0.94 & 55.64 $\pm$ 0.98 & 56.71 $\pm$ 0.92 & 56.83 $\pm$ 0.90 & 56.66 $\pm$ 0.89 & 56.83 $\pm$ 0.94 \\
        Canny + Canny & 56.29 $\pm$ 0.92 & 54.99 $\pm$ 0.96 & 58.04 $\pm$ 0.94 & 58.23 $\pm$ 0.94 & 57.95 $\pm$ 0.89 & 58.23 $\pm$ 0.93 \\
      BagNet + BagNet & 52.04 $\pm$ 0.92 & 50.34 $\pm$ 0.90 & 53.40 $\pm$ 0.98 & 53.42 $\pm$ 0.91 & 53.29 $\pm$ 0.96 & 53.42 $\pm$ 0.98 \\
      \hline
     Standard + Sobel & 53.73 $\pm$ 0.94 & 55.49 $\pm$ 0.95 & 59.01 $\pm$ 0.90 & 59.08 $\pm$ 0.91 & 58.94 $\pm$ 0.96 & 59.08 $\pm$ 0.95 \\
     Standard + Canny & 53.73 $\pm$ 1.00 & 56.29 $\pm$ 0.94 & 60.90 $\pm$ 0.94 & 60.96 $\pm$ 0.94 & 60.85 $\pm$ 0.87 & \textbf{60.96 $\pm$ 0.94} \\
    Standard + BagNet & 53.73 $\pm$ 0.95 & 52.04 $\pm$ 0.90 & 56.99 $\pm$ 0.94 & 57.17 $\pm$ 0.92 & 57.04 $\pm$ 0.91 & 57.17 $\pm$ 0.94 \\
        Sobel + Canny & 55.49 $\pm$ 0.91 & 56.29 $\pm$ 0.94 & 59.92 $\pm$ 0.95 & 60.02 $\pm$ 0.97 & 59.77 $\pm$ 0.91 & 60.02 $\pm$ 0.91 \\
       Sobel + BagNet & 55.49 $\pm$ 0.94 & 52.04 $\pm$ 0.95 & 59.17 $\pm$ 0.94 & 59.76 $\pm$ 0.96 & 59.08 $\pm$ 0.89 & 59.76 $\pm$ 0.87 \\
       Canny + BagNet & 56.29 $\pm$ 0.96 & 52.04 $\pm$ 0.95 & 61.09 $\pm$ 0.92 & 61.42 $\pm$ 0.94 & 60.68 $\pm$ 0.92 & \textbf{61.42 $\pm$ 0.93} \\
  \bottomrule
  \end{tabular}
\addtolength{\tabcolsep}{1pt}    
\end{footnotesize}

    \caption{Ensemble Baselines for STL-10}
    \end{subfigure}
    \hfill
    \caption{Full results for ensembles of pre-trained models.}
    \label{tab:big_ensemble_baselines}
\end{table}
\clearpage

\subsection{Ensembling Self-Trained Models}
\label{app:ensemble_self_train}
In Table~\ref{tab:self_trained_ensembles}, we report the best ensemble method for pairs of self-trained models
with different priors.  In Table~\ref{tab:self_trained_ensembles_big},
we report the full results over the individual ensembles. We find that, similar to the ensembles of 
models trained on the labeled data, models with diverse priors gain more from ensembling. However,
co-training models with diverse priors together still outperforms ensembling self-trained models.
\begin{table}[!h]
    \begin{subfigure}[b]{\linewidth}
        \begin{center}
            \setlength{\tabcolsep}{.7em}
\def\arraystretch{1.1}
\begin{tabular}{llccc}
    \toprule
                                     &      Feature Priors &          Model 1 &          Model 2 &         Ensemble \\
    \midrule
               \multirow{3}{*}{Same} & Standard + Standard & 59.92 $\pm$ 0.95 & 59.34 $\pm$ 0.88 & 62.25 $\pm$ 0.93 \\
                                     &       Canny + Canny & 58.40 $\pm$ 0.94 & 57.69 $\pm$ 0.94 & 60.38 $\pm$ 0.92 \\
                                     &     BagNet + BagNet & 57.80 $\pm$ 0.96 & 58.11 $\pm$ 0.85 & 60.52 $\pm$ 0.90 \\
    \hline\multirow{3}{*}{Different} &    Standard + Canny & 59.92 $\pm$ 0.90 & 58.40 $\pm$ 0.95 & \textbf{64.44 $\pm$ 0.90} \\
                                     &   Standard + BagNet & 59.92 $\pm$ 0.94 & 57.80 $\pm$ 0.96 & 63.19 $\pm$ 0.87 \\
                                     &      Canny + BagNet & 58.40 $\pm$ 0.94 & 57.80 $\pm$ 0.96 & \textbf{64.80 $\pm$ 0.91} \\
    \bottomrule
    \end{tabular}


        \end{center}
        \caption{STL-10}
    \end{subfigure}
    \begin{subfigure}[b]{\linewidth}
        \begin{center}
            \setlength{\tabcolsep}{.7em}
\def\arraystretch{1.1}
\begin{tabular}{llccc}
    \toprule
                                     &      Feature Priors &          Model 1 &          Model 2 &         Ensemble \\
    \midrule
               \multirow{3}{*}{Same} & Standard + Standard & 63.65 $\pm$ 0.81 & 61.95 $\pm$ 0.82 & 64.85 $\pm$ 0.79 \\
                                     &       Sobel + Sobel & 63.05 $\pm$ 0.81 & 66.01 $\pm$ 0.80 & 66.25 $\pm$ 0.82 \\
                                     &     BagNet + BagNet & 53.92 $\pm$ 0.82 & 52.90 $\pm$ 0.91 & 55.00 $\pm$ 0.83 \\
    \hline\multirow{3}{*}{Different} &    Standard + Sobel & 63.65 $\pm$ 0.81 & 63.05 $\pm$ 0.83 & \textbf{67.52 $\pm$ 0.77} \\
                                     &   Standard + BagNet & 63.65 $\pm$ 0.81 & 53.92 $\pm$ 0.88 & 64.10 $\pm$ 0.79 \\
                                     &      Sobel + BagNet & 63.05 $\pm$ 0.83 & 53.92 $\pm$ 0.89 & 65.68 $\pm$ 0.79 \\
    \bottomrule
    \end{tabular}


        \end{center}
        \caption{CIFAR-10}
    \end{subfigure}
    \caption{Ensemble performance when combining \textit{self-trained} models with Standard, Canny, Sobel, and BagNet priors. 
    When two models of the same prior are ensembled, the models are trained with different 
    random initializations.}
    \label{tab:self_trained_ensembles}
\end{table}

\begin{table}[!h]
    \begin{subfigure}[b]{\textwidth}
        \centering
        \begin{footnotesize}
    \addtolength{\tabcolsep}{-1pt}   
    \begin{tabular}{lccccccc}
        \toprule
             Feature Priors &          Model 1 &          Model 2 &   Max Conf. &    Avg Conf. &        Rank &        Best \\
        \midrule
        Standard + Standard & 63.65 $\pm$ 0.81 & 61.95 $\pm$ 0.87 & 64.84 $\pm$ 0.77 & 64.85 $\pm$ 0.76 & 64.83 $\pm$ 0.83 & 64.85 $\pm$ 0.79 \\
              Sobel + Sobel & 63.05 $\pm$ 0.87 & 66.01 $\pm$ 0.82 & 66.19 $\pm$ 0.81 & 66.25 $\pm$ 0.79 & 66.17 $\pm$ 0.81 & 66.25 $\pm$ 0.83 \\
            BagNet + BagNet & 53.92 $\pm$ 0.87 & 52.90 $\pm$ 0.83 & 54.86 $\pm$ 0.87 & 55.00 $\pm$ 0.83 & 54.87 $\pm$ 0.82 & 55.00 $\pm$ 0.87 \\
            \hline
           Standard + Sobel & 63.65 $\pm$ 0.79 & 63.05 $\pm$ 0.80 & 67.42 $\pm$ 0.79 & 67.52 $\pm$ 0.79 & 67.38 $\pm$ 0.79 & \textbf{67.52 $\pm$ 0.77} \\
           Standard + Canny & 63.65 $\pm$ 0.90 & 51.82 $\pm$ 0.88 & 63.70 $\pm$ 0.81 & 63.91 $\pm$ 0.81 & 63.02 $\pm$ 0.83 & 63.91 $\pm$ 0.82 \\
          Standard + BagNet & 63.65 $\pm$ 0.81 & 53.92 $\pm$ 0.82 & 64.05 $\pm$ 0.85 & 64.10 $\pm$ 0.79 & 62.69 $\pm$ 0.80 & 64.10 $\pm$ 0.86 \\
              Sobel + Canny & 63.05 $\pm$ 0.81 & 51.82 $\pm$ 0.80 & 61.43 $\pm$ 0.80 & 61.42 $\pm$ 0.80 & 60.66 $\pm$ 0.81 & 61.43 $\pm$ 0.83 \\
             Sobel + BagNet & 63.05 $\pm$ 0.78 & 53.92 $\pm$ 0.83 & 65.45 $\pm$ 0.85 & 65.68 $\pm$ 0.82 & 64.65 $\pm$ 0.80 & 65.68 $\pm$ 0.82 \\
             Canny + BagNet & 51.82 $\pm$ 0.81 & 53.92 $\pm$ 0.79 & 59.60 $\pm$ 0.81 & 59.79 $\pm$ 0.83 & 60.24 $\pm$ 0.82 & 60.24 $\pm$ 0.81 \\
        \bottomrule
        \end{tabular}
              
    \addtolength{\tabcolsep}{1pt}    
\end{footnotesize}

    \caption{Ensemble Baselines for CIFAR-10}
    \end{subfigure}

    \begin{subfigure}[b]{\textwidth}
        \centering
        \begin{footnotesize}
    \addtolength{\tabcolsep}{-1pt}    
    \begin{tabular}{lccccccc}
        \toprule
             Feature Priors &          Model 1 &          Model 2 &   Max Conf. &    Avg Conf. &        Rank &        Best \\
        \midrule
        Standard + Standard & 59.92 $\pm$ 0.92 & 59.34 $\pm$ 0.99 &  62.18 $\pm$ 0.92 & 62.25 $\pm$ 0.96 & 62.16 $\pm$ 0.88 & 62.25 $\pm$ 0.94 \\
              Canny + Canny & 58.40 $\pm$ 0.95 & 57.69 $\pm$ 0.89 &  60.30 $\pm$ 0.95 & 60.36 $\pm$ 0.92 & 60.38 $\pm$ 0.91 & 60.38 $\pm$ 0.95 \\
            BagNet + BagNet & 57.80 $\pm$ 0.89 & 58.11 $\pm$ 0.94 &  60.42 $\pm$ 0.90 & 60.46 $\pm$ 0.98 & 60.52 $\pm$ 0.93 & 60.52 $\pm$ 0.90 \\
            \hline
           Standard + Sobel & 59.92 $\pm$ 0.92 & 57.86 $\pm$ 0.91 &  62.49 $\pm$ 0.89 & 62.69 $\pm$ 0.91 & 62.66 $\pm$ 0.89 & 62.69 $\pm$ 0.94 \\
           Standard + Canny & 59.92 $\pm$ 0.94 & 58.40 $\pm$ 0.95 &  64.29 $\pm$ 0.95 & 64.44 $\pm$ 0.89 & 64.34 $\pm$ 0.95 & \textbf{64.44 $\pm$ 0.95} \\
          Standard + BagNet & 59.92 $\pm$ 0.89 & 57.80 $\pm$ 0.97 &  63.01 $\pm$ 0.93 & 63.10 $\pm$ 0.89 & 63.19 $\pm$ 0.88 & 63.19 $\pm$ 0.88 \\
              Sobel + Canny & 57.86 $\pm$ 0.91 & 58.40 $\pm$ 0.93 &  62.20 $\pm$ 0.92 & 62.14 $\pm$ 0.92 & 62.22 $\pm$ 0.90 & 62.22 $\pm$ 0.91 \\
             Sobel + BagNet & 57.86 $\pm$ 0.95 & 57.80 $\pm$ 0.95 &  62.24 $\pm$ 0.94 & 62.58 $\pm$ 0.90 & 63.52 $\pm$ 0.91 & 63.52 $\pm$ 0.88 \\
             Canny + BagNet & 58.40 $\pm$ 0.93 & 57.80 $\pm$ 0.95 &  64.38 $\pm$ 0.89 & 64.64 $\pm$ 0.92 & 64.80 $\pm$ 0.90 & \textbf{64.80 $\pm$ 0.92} \\
        \bottomrule
        \end{tabular}
        
    \addtolength{\tabcolsep}{1pt}    
\end{footnotesize}

    \caption{Ensemble Baselines for STL-10}
    \end{subfigure}
    \caption{Full results for ensembles of self-trained models.}
    \label{tab:self_trained_ensembles_big}
\end{table}

\clearpage
\subsection{Stacked Ensembling}
\label{app:stacked_ensembling}
Here we consider an ensembling technique that leverages a validation set. We implement stacking (also called blending) 
\cite{toscher2009bigchaos, sill2009feature}, 
which takes in the outputs of the member models as input, and then trains a second model to produce the final 
layer. Here, we take the logits of each model in the ensemble, and train the secondary model using logistic regression 
on the validation set for the dataset. We report accuracies of the ensemble on the test set below. We again find that 
prior diversity is important for the performance of the ensemble. 
\begin{table}[!h]
    \centering
    \begin{footnotesize}
    \def\arraystretch{1.1}
      \addtolength{\tabcolsep}{-1pt}    
      \begin{tabular}{l|ccc|ccc}
        \toprule
        & \multicolumn{3}{c|}{Pre-trained} & \multicolumn{3}{c}{Self-trained} \\
        \midrule 
             Feature Priors &   Model 1 &  Model 2 & \begin{tabular}{c} Stacked \\ Ensemble \end{tabular} & Model 1 &  Model 2 &  \begin{tabular}{c} Stacked \\ Ensemble \end{tabular}\\
        \midrule
        Standard + Standard & 52.54 $\pm$ 0.85 & 51.82 $\pm$ 0.85 & 54.13 $\pm$ 0.88 & 63.65 $\pm$ 0.81 & 61.95 $\pm$ 0.82 & 65.13 $\pm$ 0.82\\
              Sobel + Sobel & 51.94 $\pm$ 0.88 & 53.69 $\pm$ 0.86 & 54.46 $\pm$ 0.92 & 63.05 $\pm$ 0.81 & 66.01 $\pm$ 0.80 & 66.35 $\pm$ 0.80\\
            BagNet + BagNet & 42.22 $\pm$ 0.80 & 42.56 $\pm$ 0.83 & 44.28 $\pm$ 0.83 & 53.92 $\pm$ 0.82 & 52.90 $\pm$ 0.91 & 54.94 $\pm$ 0.84\\
            \hline
           Standard + Sobel & 52.54 $\pm$ 0.79 & 51.94 $\pm$ 0.82 & \textbf{57.42 $\pm$ 0.84} & 63.65 $\pm$ 0.81 & 63.05 $\pm$ 0.83 & \textbf{67.01 $\pm$ 0.79} \\
          Standard + BagNet & 52.54 $\pm$ 0.85 & 42.22 $\pm$ 0.80 & 53.65 $\pm$ 0.85          & 63.65 $\pm$ 0.81 & 53.92 $\pm$ 0.88 & 64.61 $\pm$ 0.81 \\
             Sobel + BagNet & 51.94 $\pm$ 0.86 & 42.22 $\pm$ 0.82 & 55.75 $\pm$ 0.83          & 63.05 $\pm$ 0.83 & 53.92 $\pm$ 0.89 & 65.67 $\pm$ 0.82 \\
        \bottomrule
        \end{tabular}
        
        \addtolength{\tabcolsep}{1pt}    
    \end{footnotesize}
    \vspace{1em}
    \caption{Performance of ensembling pre-trained and self-trained models with
    stacked ensembling on CIFAR-10.}
\end{table}

\begin{table}[!h]
    \centering
    \begin{footnotesize}
    \def\arraystretch{1.1}
      \addtolength{\tabcolsep}{-1pt}    
      \begin{tabular}{l|ccc|ccc}
        \toprule
        & \multicolumn{3}{c|}{Pre-trained} & \multicolumn{3}{c}{Self-trained} \\
        \midrule 
             Feature Priors &   Model 1 &  Model 2 & \begin{tabular}{c} Stacked \\ Ensemble \end{tabular} & Model 1 &  Model 2 &  \begin{tabular}{c} Stacked \\ Ensemble \end{tabular}\\
        \midrule
        Standard + Standard & 53.73 $\pm$ 0.86 & 55.38 $\pm$ 1.00 & 56.01 $\pm$ 0.94             & 59.92 $\pm$ 0.95 & 59.34 $\pm$ 0.88 & 60.54 $\pm$ 0.91\\
              Canny + Canny & 56.29 $\pm$ 0.92 & 54.99 $\pm$ 0.96 & 57.70 $\pm$ 0.90             & 58.40 $\pm$ 0.94 & 57.69 $\pm$ 0.94 & 59.23 $\pm$ 0.99\\
            BagNet + BagNet & 52.04 $\pm$ 0.92 & 50.34 $\pm$ 0.90 & 52.35 $\pm$ 0.97             & 57.80 $\pm$ 0.96 & 58.11 $\pm$ 0.85 & 59.48 $\pm$ 0.98\\
            \hline  
           Standard + Canny & 53.73 $\pm$ 1.00 & 56.29 $\pm$ 0.94 & 59.24 $\pm$ 0.88             & 59.92 $\pm$ 0.90 & 58.40 $\pm$ 0.95 & \textbf{63.42 $\pm$ 0.89}\\
          Standard + BagNet & 53.73 $\pm$ 0.95 & 52.04 $\pm$ 0.90 & 56.03 $\pm$ 0.98             & 59.92 $\pm$ 0.94 & 57.80 $\pm$ 0.96 & 62.59 $\pm$ 0.91\\
             Canny + BagNet & 56.29 $\pm$ 0.96 & 52.04 $\pm$ 0.95 & \textbf{59.98 $\pm$ 0.91}    & 58.40 $\pm$ 0.94 & 57.80 $\pm$ 0.96 & 63.22 $\pm$ 0.94\\
        \bottomrule
        \end{tabular}
        
        \addtolength{\tabcolsep}{1pt}    
    \end{footnotesize}

    \vspace{1em}
    \caption{Performance of ensembling pre-trained and self-trained models with
    stacked ensembling on STL-10.}
\end{table}

\clearpage
\subsection{Self-Training and Co-Training on STL-10 and CIFAR-10}
\label{app:big_basic_table}
\begin{table}[!h]
    \centering
    \begin{tabular}{llccc}
    \toprule
                                   Methods &  Prior(s) &     Labeled Only & \begin{tabular}{c}+Unlabeled \\Self/Co-Training\end{tabular} & \begin{tabular}{c} + Standard model \\with Pseudo-labels \end{tabular} \\
    \midrule
            \multirow{4}{*}{Self-training} &  Standard & 52.54 $\pm$ 0.86 &                                   63.65 $\pm$ 0.76 &                                   64.02 $\pm$ 0.82 \\
                                           &     Canny & 45.48 $\pm$ 0.90 &                                   51.82 $\pm$ 0.82 &                                   55.59 $\pm$ 0.80 \\
                                           &     Sobel & 51.94 $\pm$ 0.88 &                                   63.05 $\pm$ 0.84 &                                   64.77 $\pm$ 0.80 \\
                                           &    BagNet & 42.22 $\pm$ 0.82 &                                   53.92 $\pm$ 0.89 &                                   54.21 $\pm$ 0.85 \\
    \midrule \multirow{20}{*}{Co-training} &  Standard & 52.54 $\pm$ 0.91 &                                   65.06 $\pm$ 0.76 &                  \multirow{2}{*}{65.10 $\pm$ 0.84} \\
                                           & +Standard & 51.82 $\pm$ 0.86 &                                   64.93 $\pm$ 0.80 &                                                    \\
                               \cline{2-5} &     Canny & 45.48 $\pm$ 0.85 &                                   51.15 $\pm$ 0.79 &                  \multirow{2}{*}{55.74 $\pm$ 0.80} \\
                                           &    +Canny & 44.19 $\pm$ 0.82 &                                   51.65 $\pm$ 0.81 &                                                    \\
                               \cline{2-5} &     Sobel & 51.94 $\pm$ 0.86 &                                   67.18 $\pm$ 0.80 &                  \multirow{2}{*}{68.47 $\pm$ 0.74} \\
                                           &    +Sobel & 53.69 $\pm$ 0.89 &                                   67.35 $\pm$ 0.77 &                                                    \\
                               \cline{2-5} &     Canny & 45.48 $\pm$ 0.79 &                                   58.66 $\pm$ 0.81 &                  \multirow{2}{*}{65.34 $\pm$ 0.81} \\
                                           &    +Sobel & 51.94 $\pm$ 0.80 &                                   64.87 $\pm$ 0.79 &                                                    \\
                               \cline{2-5} &     Canny & 45.48 $\pm$ 0.85 &                                   59.19 $\pm$ 0.85 &                  \multirow{2}{*}{67.59 $\pm$ 0.74} \\
                                           &   +BagNet & 42.22 $\pm$ 0.85 &                                   67.92 $\pm$ 0.79 &                                                    \\
                               \cline{2-5} &     Sobel & 51.94 $\pm$ 0.81 & 71.88 $\pm$ 0.73 & \multirow{2}{*}{\textbf{74.25 $\pm$ 0.74}} \\
                                           &   +BagNet & 42.22 $\pm$ 0.82 &                                   73.91 $\pm$ 0.71 &                                                    \\
                               \cline{2-5} &    BagNet & 42.22 $\pm$ 0.79 &                                   55.94 $\pm$ 0.83 &                  \multirow{2}{*}{56.05 $\pm$ 0.77} \\
                                           &   +BagNet & 42.56 $\pm$ 0.86 &                                   55.26 $\pm$ 0.88 &                                                    \\
                               \cline{2-5} &     Canny & 45.48 $\pm$ 0.85 &                                   59.23 $\pm$ 0.81 &                  \multirow{2}{*}{67.21 $\pm$ 0.77} \\
                                           & +Standard & 52.54 $\pm$ 0.87 &                                   66.92 $\pm$ 0.82 &                                                    \\
                               \cline{2-5} &     Sobel & 51.94 $\pm$ 0.83 & 71.44 $\pm$ 0.76 & \multirow{2}{*}{\textbf{73.83 $\pm$ 0.76}} \\
                                           & +Standard & 52.54 $\pm$ 0.85 &                                   73.59 $\pm$ 0.72 &                                                    \\
                               \cline{2-5} &  Standard & 52.54 $\pm$ 0.88 &                                   66.67 $\pm$ 0.83 &                  \multirow{2}{*}{66.77 $\pm$ 0.75} \\
                                           &   +BagNet & 42.22 $\pm$ 0.80 &                                   67.12 $\pm$ 0.75 &                                                    \\
    \bottomrule
\end{tabular}

    \vspace{1em}
    \caption{Performance of self-training and co-training on CIFAR-10
    for each prior combination. }
    \label{tab:big_basic_table_cifar}
\end{table}
\begin{table}[!h]
    \centering
        \begin{tabular}{llccc}
    \toprule
                                   Methods &  Prior(s) &     Labeled Only & \begin{tabular}{c}+Unlabeled \\Self/Co-Training\end{tabular} & \begin{tabular}{c} + Standard model \\with Pseudo-labels \end{tabular} \\
    \midrule
            \multirow{4}{*}{Self-training} &  Standard & 53.73 $\pm$ 0.95 &                                   59.92 $\pm$ 0.91 &                                   60.52 $\pm$ 0.94 \\
                                           &     Canny & 56.29 $\pm$ 0.96 &                                   58.40 $\pm$ 0.91 &                                   62.19 $\pm$ 0.92 \\
                                           &     Sobel & 55.49 $\pm$ 0.96 &                                   57.86 $\pm$ 0.98 &                                   60.92 $\pm$ 0.89 \\
                                           &    BagNet & 52.04 $\pm$ 0.96 &                                   57.80 $\pm$ 0.99 &                                   61.69 $\pm$ 0.95 \\
    \midrule \multirow{20}{*}{Co-training} &  Standard & 53.73 $\pm$ 0.95 &                                   58.05 $\pm$ 0.92 &                  \multirow{2}{*}{61.16 $\pm$ 0.95} \\
                                           & +Standard & 55.38 $\pm$ 0.96 &                                   60.44 $\pm$ 0.95 &                                                    \\
                               \cline{2-5} &     Canny & 56.29 $\pm$ 0.92 &                                   60.22 $\pm$ 0.91 &                  \multirow{2}{*}{63.24 $\pm$ 0.92} \\
                                           &    +Canny & 54.99 $\pm$ 0.94 &                                   59.56 $\pm$ 0.94 &                                                    \\
                               \cline{2-5} &     Sobel & 55.49 $\pm$ 0.96 &                                   58.93 $\pm$ 0.91 &                  \multirow{2}{*}{60.68 $\pm$ 0.94} \\
                                           &    +Sobel & 55.64 $\pm$ 0.95 &                                   59.23 $\pm$ 0.90 &                                                    \\
                               \cline{2-5} &     Canny & 56.29 $\pm$ 0.95 &                                   62.40 $\pm$ 0.99 &                  \multirow{2}{*}{65.53 $\pm$ 0.84} \\
                                           &    +Sobel & 55.49 $\pm$ 0.92 &                                   64.11 $\pm$ 0.91 &                                                    \\
                               \cline{2-5} &     Canny & 56.29 $\pm$ 0.92 & 62.21 $\pm$ 0.89 & \multirow{2}{*}{\textbf{67.33 $\pm$ 0.88}} \\
                                           &   +BagNet & 52.04 $\pm$ 0.94 &                                   66.74 $\pm$ 0.87 &                                                    \\
                               \cline{2-5} &     Sobel & 55.49 $\pm$ 0.92 &                                   62.72 $\pm$ 0.94 &                  \multirow{2}{*}{65.79 $\pm$ 0.94} \\
                                           &   +BagNet & 52.04 $\pm$ 1.00 &                                   65.44 $\pm$ 0.91 &                                                    \\
                               \cline{2-5} &    BagNet & 52.04 $\pm$ 0.89 &                                   59.85 $\pm$ 0.89 &                  \multirow{2}{*}{60.84 $\pm$ 0.95} \\
                                           &   +BagNet & 50.34 $\pm$ 0.91 &                                   60.16 $\pm$ 0.89 &                                                    \\
                               \cline{2-5} &     Canny & 56.29 $\pm$ 0.94 &                                   62.16 $\pm$ 0.92 &                  \multirow{2}{*}{65.67 $\pm$ 0.93} \\
                                           & +Standard & 53.73 $\pm$ 0.92 &                                   64.22 $\pm$ 0.91 &                                                    \\
                               \cline{2-5} &     Sobel & 55.49 $\pm$ 0.95 &                                   61.15 $\pm$ 0.89 &                  \multirow{2}{*}{63.08 $\pm$ 0.91} \\
                                           & +Standard & 53.73 $\pm$ 0.92 &                                   61.74 $\pm$ 0.93 &                                                    \\
                               \cline{2-5} &  Standard & 53.73 $\pm$ 0.94 &                                   61.99 $\pm$ 0.88 &                  \multirow{2}{*}{62.34 $\pm$ 0.89} \\
                                           &   +BagNet & 52.04 $\pm$ 0.91 &                                   62.31 $\pm$ 1.00 &                                                    \\
    \bottomrule
    \end{tabular}

    \vspace{1em}
    \caption{Performance of self-training and co-training on STL-10
    for each prior combination. }
    \label{tab:big_basic_table_stl}
\end{table}

\clearpage
\subsection{Additional Results for Ensembling Diverse Feature Priors (Full CIFAR-10, ImageNet)}
\label{app:full_cifar10_IN}
In Figures~\ref{app_fig:cifarfull} and~\ref{app_fig:imagenet}, we perform an analysis of using an 
ensemble to combine models trained on the full CIFAR-10 and the ImageNet (96x96) dataset
respectively. We find that models with different feature priors still have 
less correlated predictions than those of the same feature prior, and 
thus have less overlapping failure modes. Ensembles of models with diverse 
priors provide a significant boost over the performance of individual models, 
higher than that of combining models trained with the same prior. 
It is worth noting that, in these settings, the specific feature priors we
introduce result in models with accuracy significantly lower than that of a
standard model. Designing better domain-specific priors is thus an important avenue for future work.

\begin{table}[!htbp]
    \centering
    \begin{subtable}[c]{0.5\textwidth}
    \centering
    \setlength{\tabcolsep}{.4em}
    \def\arraystretch{1.1}
    
    \begin{tabular}{l|cccc}
        \toprule
                 & Standard & Sobel & Canny & BagNet \\
        \midrule
        Standard & 0.583 & 0.369 & 0.242 & 0.473\\   
        Sobel    &       & 0.597 & 0.358 & 0.295 \\
        Canny    &       &       & 0.594 & 0.212\\
        BagNet   &       &       &       & 0.594\\
        \bottomrule
    \end{tabular}
    \caption{Correlation of correct predictions (cf. Table~\ref{tab:independence})
    \vspace{.3cm}}
    \end{subtable}
    \begin{subtable}[c]{0.75\textwidth}
    \centering
    \setlength{\tabcolsep}{.4em}
    \def\arraystretch{1.1}
    \begin{tabular}{lccc}
        \toprule
        Feature Priors &          Model 1 &          Model 2 &         Ensemble \\
        \midrule
        Standard + Standard & 91.73 $\pm$ 0.44 & 91.97 $\pm$ 0.46 & 92.83 $\pm$ 0.44 \\
        Sobel + Sobel & 86.18 $\pm$ 0.58 & 86.21 $\pm$ 0.58 & 87.43 $\pm$ 0.59 \\
        BagNet + BagNet & 90.47 $\pm$ 0.49 & 90.85 $\pm$ 0.49 & 91.69 $\pm$ 0.48 \\
        \hline 
        Standard + Sobel & 91.73 $\pm$ 0.44 & 86.18 $\pm$ 0.58 & 92.23 $\pm$ 0.44  \\
        Standard + BagNet & 91.73 $\pm$ 0.44 & 90.47 $\pm$ 0.49 & 93.01 $\pm$ 0.42 \\
        Sobel + BagNet & 86.18 $\pm$ 0.58 & 90.47 $\pm$ 0.49 & 92.27 $\pm$ 0.44\\
        \bottomrule
    \end{tabular}
    \caption{Ensemble accuracy (cf. Table~\ref{tab:ensembles}).}
    \end{subtable}
    \caption{Full CIFAR-10 dataset.}
    \label{app_fig:cifarfull}
    \end{table}
    \vspace{-1em}

    \begin{table}[!ht]
    \centering
        \begin{subtable}[c]{0.5\textwidth}
            \centering
            \setlength{\tabcolsep}{.4em}
            \def\arraystretch{1.1}
    
    \begin{tabular}{l|cccc}
        \toprule
                 & Standard & Sobel  & BagNet \\
        \midrule
        Standard & 0.6528&  0.4925 & 0.5613 \\   
        Sobel    &       &  0.6384 & 0.4529 \\
        BagNet   &       &         &  0.6517 \\
        \bottomrule
    \end{tabular}
    \caption{Correlation of correct predictions (cf. Table~\ref{tab:independence})
    \vspace{.3cm}}
    \end{subtable}\\

    \begin{subtable}[c]{0.75\textwidth}
        \centering
        \setlength{\tabcolsep}{.4em}
        \def\arraystretch{1.1}
    \begin{tabular}{lccc}
        \toprule
        Feature Priors &          Model 1 &          Model 2 &         Ensemble \\
        \midrule
        Standard + Standard &  60.34 $\pm$ 0.38 & 60.30 $\pm$ 0.38 & 63.36 $\pm$ 0.36\\
        Sobel + Sobel & 51.87 $\pm$ 0.36 & 51.62 $\pm$ 0.39 & 54.90 $\pm$ 0.35 \\
        BagNet + BagNet &  52.57 $\pm$ 0.36 & 52.38 $\pm$ 0.38 & 55.42 $\pm$ 0.38 \\
        \hline 
        Standard + Sobel & 60.34 $\pm$ 0.37 & 51.87 $\pm$ 0.36 & 62.86 $\pm$ 0.37\\
        Standard + BagNet & 60.34 $\pm$ 0.38 & 52.57 $\pm$ 0.36 & 62.00 $\pm$ 0.36\\
        Sobel + BagNet & 51.87 $\pm$ 0.36 & 52.57 $\pm$ 0.36 & 59.41 $\pm$ 0.36\\
        \bottomrule
    \end{tabular}
    \caption{Ensemble accuracy (cf. Table~\ref{tab:ensembles}).}
    \end{subtable}
    \caption{ImageNet dataset.}
    \label{app_fig:imagenet}
    \end{table}

\newpage
\subsection{Co-Training with varying amounts of labeled data.}
\label{app:vary_label_data_stl10}
In Table \ref{tab:changing_stl_labels}, we study how the efficacy of combining diverse priors through cotraining changes 
as the number of labeled examples increase for STL-10. As one might expect, when labeled data is sparse, 
the feature priors learned by the models alone are relatively brittle: thus, leveraging 
diverse priors against each other on unlabeled data improves generalization. 
As the number of labeled examples increases, the models with single feature priors learn more reliable prediction rules that 
can already generalize, so the additional benefit of combining feature priors diminishes. However, even in settings with 
plentiful data, combining diverse feature priors can aid generalization if there is a spurious correlation in the labeled data (see Section \ref{sec:spurious}.)

\begin{table}[!h]
    \centering
    \begin{tabular}{ccc}
    \toprule
    Number of Labeled Examples &  Standard + Standard &    Canny + BagNet\\
    \midrule
    1000 & 61.16 $\pm$ 0.94 & \textbf{67.33 $\pm$ 0.89 }\\
    2000 & 68.24 $\pm$ 1.12 & \textbf{72.76 $\pm$ 1.08}\\
    3000 & 74.88 $\pm$ 0.97 & \textbf{75.76 $\pm$ 1.04}\\
    4000 & \textbf{78.85} $\pm$ 0.99 & 77.44 $\pm$ 1.00\\
    \bottomrule
\end{tabular}

\vspace{1em}
\caption{Performance of co-training with varying amounts of training data for STL-10. }
\label{tab:changing_stl_labels}
\end{table}

\clearpage
\subsection{Correlation between the individual feature-biased models and the
final standard model}
\label{app:independence_after_cotraining}
\begin{table}[!h]
    \centering
        
\setlength{\tabcolsep}{1.5em}
\begin{tabular}{ll|cc|cc}
    \toprule
    & & \multicolumn{2}{c|}{CIFAR-10} & \multicolumn{2}{c}{STL-10}\\
    Method & Prior  & Before &  After & Before &  After \\
    \midrule
    \multirow{4}{*}{Self-training} & Standard &        0.598 &       0.813 &        0.554 &       0.728 \\
                                   &    Canny &        0.237 &       0.622 &        0.305 &       0.519 \\
                                   &    Sobel &        0.259 &        0.76 &        0.385 &       0.621 \\
                                   &   BagNet &         0.38 &       0.752 &        0.357 &       0.516 \\
\hline\multirow{4}{*}{Co-training} &    Canny &        0.237 &       0.595 &        0.305 &       0.496 \\
                                   &  +BagNet &         0.38 &       0.664 &        0.357 &       0.538 \\
                       \cline{2-6} &    Sobel &        0.259 &       0.719 &        0.385 &       0.581 \\
                                   &  +BagNet &         0.38 &       0.716 &        0.357 &       0.554 \\
\bottomrule    
\end{tabular}

     

        \vspace{1em}
    \caption{Similarity between models
        before and after training on pseudo-labeled data.
        Our measure of similarity is the (Pearson) correlation between which
        test examples are correctly predicted by each model.
        In Columns 3 and 5 we report that notion of similarity between the
        pre-trained feature-biased models and the pre-trained standard model
        (the numbers are reproduced from Table~\ref{tab:independence}).
        Then, in columns 4 and 6 we report the similarity between the
        feature-biased models at the end of self- or co-training and the
        standard model trained on their (potentially combined) pseudo-labels.
        We observe that through this process of training a standard model on the
        pseudo-labels of different feature-biased models, the former behaves
        more similar to the latter.
}
    \label{tab:independence_after_cotraining}
\end{table}

\clearpage
\subsection{Ensembles for Spurious Datasets}
\label{app:spurious_ensembles}
In Table~\ref{tab:spurious_ensembles} (full table in Table~\ref{tab:spurious_ensembles_big}), we ensemble the 
self-trained priors for the Tinted STL-10 dataset and the CelebA dataset as in Section~\ref{sec:spurious}. 
Both of these datasets have a spurious correlation base on color, which results in a weak Standard and BagNet model.
As a result, the ensembles with the Standard or BagNet models do not perform well on the test set. However, 
in Section~\ref{tab:spurioussynthetic}, we find that co-training in this setting allows the BagNet model to improve 
when jointly trained with a shape model, thus boosting the final performance.  
\begin{table}[!h]
    \centering
    \begin{subfigure}[b]{\linewidth}
        \centering
        \begin{tabular}{lccc}
    \toprule
    Feature Priors &          Model 1 &          Model 2 &         Ensemble \\
    \midrule
    Standard + Canny & 17.56 $\pm$ 0.73 & \textbf{57.31 $\pm$ 0.96} & 44.31 $\pm$ 0.90 \\
    Standard + Sobel & 17.56 $\pm$ 0.71 & 56.12 $\pm$ 0.90 & 46.06 $\pm$ 0.95 \\
    Standard + BagNet & 17.56 $\pm$ 0.73 & 13.53 $\pm$ 0.66 & 16.64 $\pm$ 0.66 \\
    Canny + BagNet & \textbf{57.31 $\pm$ 0.96} & 13.53 $\pm$ 0.64 & 48.30 $\pm$ 0.89 \\
    Sobel + BagNet & 56.12 $\pm$ 0.91 & 13.53 $\pm$ 0.69 & 49.05 $\pm$ 0.98 \\
    \bottomrule
\end{tabular}

    \caption{Tinted STL-10}
    \end{subfigure}
    \hfill

    \begin{subfigure}[b]{\textwidth}
        \centering
        \begin{tabular}{lccc}
    \toprule
    Feature Priors &          Model 1 &          Model 2 &         Ensemble \\
    \midrule
    Standard + Canny & 71.57 $\pm$ 0.53 & \textbf{85.73 $\pm$ 0.40} & 84.05 $\pm$ 0.42 \\
        Standard + Sobel & 71.57 $\pm$ 0.55 & \textbf{85.42 $\pm$ 0.43} & 82.10 $\pm$ 0.45 \\
    Standard + BagNet & 71.57 $\pm$ 0.53 & 64.89 $\pm$ 0.56 & 69.66 $\pm$ 0.55 \\
    Canny + BagNet & \textbf{85.73 $\pm$ 0.42} & 64.89 $\pm$ 0.56 & 84.06 $\pm$ 0.45 \\
    Sobel + BagNet & \textbf{85.42 $\pm$ 0.43} & 64.89 $\pm$ 0.57 & 82.89 $\pm$ 0.44 \\
    \bottomrule
    \end{tabular}

    \caption{CelebA}
    \end{subfigure}
    \caption{Performance of ensembles consisting of models trained with
    different priors.}
    \label{tab:spurious_ensembles}
\end{table}

\begin{table}[!h]
    \hfill
    \begin{subfigure}[b]{\textwidth}
        \centering
        \begin{footnotesize}
    \addtolength{\tabcolsep}{-1pt}  
\begin{tabular}{lccccccc}
    \toprule
       Feature Priors &          Model 1 &          Model 2 &   Max Conf. &    Avg Conf. &        Rank &        Best \\
    \midrule
       Standard + Canny & 17.56 $\pm$ 0.70 & \textbf{57.31 $\pm$ 0.95} & 44.31 $\pm$ 0.98 & 43.48 $\pm$ 0.94 & 42.12 $\pm$ 0.95 & 44.31 $\pm$ 0.94 \\
     Standard + Sobel & 17.56 $\pm$ 0.66 & 56.12 $\pm$ 0.98 & 46.06 $\pm$ 0.94 & 44.71 $\pm$ 0.91 & 39.39 $\pm$ 0.95 & 46.06 $\pm$ 0.99 \\
    Standard + BagNet & 17.56 $\pm$ 0.71 & 13.53 $\pm$ 0.64 & 16.59 $\pm$ 0.69 & 16.64 $\pm$ 0.71 & 16.14 $\pm$ 0.74 & 16.64 $\pm$ 0.66 \\
    Canny + BagNet & \textbf{57.31 $\pm$ 0.91} & 13.53 $\pm$ 0.62 & 48.09 $\pm$ 0.96 & 48.30 $\pm$ 1.01 & 39.92 $\pm$ 0.92 & 48.30 $\pm$ 0.95 \\
       Sobel + BagNet & 56.12 $\pm$ 0.94 & 13.53 $\pm$ 0.64 & 49.00 $\pm$ 0.95 & 49.05 $\pm$ 0.95 & 37.67 $\pm$ 0.91 & 49.05 $\pm$ 0.93 \\
    \bottomrule
    \end{tabular}
    \addtolength{\tabcolsep}{1pt}    
\end{footnotesize}

    \caption{Tinted STL-10}
    \end{subfigure}

    \begin{subfigure}[b]{\textwidth}
        \centering
        \begin{footnotesize}
    \addtolength{\tabcolsep}{-1pt}  
\begin{tabular}{lccccccc}
    \toprule
       Feature Priors &          Model 1 &          Model 2 &   Max Conf. &    Avg Conf. &        Rank &        Best \\
    \midrule
       Standard + Canny & 71.57 $\pm$ 0.53 & \textbf{85.73 $\pm$ 0.43} & 83.96 $\pm$ 0.44 & 84.05 $\pm$ 0.43 & 84.00 $\pm$ 0.46 & 84.05 $\pm$ 0.43 \\
       Standard + Sobel & 71.57 $\pm$ 0.57 & \textbf{85.42 $\pm$ 0.41} & 82.06 $\pm$ 0.45 & 82.10 $\pm$ 0.45 & 78.01 $\pm$ 0.51 & 82.10 $\pm$ 0.49 \\
    Standard + BagNet & 71.57 $\pm$ 0.56 & 64.89 $\pm$ 0.56 & 69.66 $\pm$ 0.54 & 69.66 $\pm$ 0.54 & 68.01 $\pm$ 0.58 & 69.66 $\pm$ 0.54 \\
    Canny + BagNet & \textbf{85.73 $\pm$ 0.42} & 64.89 $\pm$ 0.57 & 84.06 $\pm$ 0.44 & 84.06 $\pm$ 0.45 & 72.79 $\pm$ 0.51 & 84.06 $\pm$ 0.44 \\
    Sobel + BagNet & \textbf{85.42 $\pm$ 0.39} & 64.89 $\pm$ 0.55 & 82.89 $\pm$ 0.46 & 82.89 $\pm$ 0.46 & 71.65 $\pm$ 0.57 & 82.89 $\pm$ 0.43 \\
    \bottomrule
    \end{tabular}
    \addtolength{\tabcolsep}{1pt}    
\end{footnotesize}

    \caption{CelebA}
    \end{subfigure}
    \hfill
    \caption{Performance of individual ensembles on datasets with spurious
    correlations.}
    \label{tab:spurious_ensembles_big}    
\end{table}

\clearpage
\subsection{Breakdown of test accuracy for co-training on CelebA}
\label{app:splitceleba}
\begin{table}[!h]
    \begin{center}
        \begin{tabular}{lccccc}
    \toprule
                                   Method &                                        Prior(s) & \begin{tabular}{c} Female \\ Blond \\ (N=2480)\end{tabular} & \begin{tabular}{c} Female \\ Not Blond \\ (N=9767)\end{tabular} & \begin{tabular}{c} Male \\ Blond \\ (N=180)\end{tabular} & \begin{tabular}{c} Male \\ Not Blond \\ (N=7535)\end{tabular} \\
    \midrule
           \multirow{4}{*}{Self-training} &
           Standard &                                   \textbf{97.78 $\pm$
           0.52} &                                   47.06 $\pm$ 0.83 &                                   55.56 $\pm$ 6.11 &                                   95.94 $\pm$ 0.37 \\
                                          &
           Canny &                                   94.44 $\pm$ 0.81 &
           77.27 $\pm$ 0.69 &                                   \textbf{78.33
           $\pm$ 5.00} &                                   96.19 $\pm$ 0.36 \\
                                          &                                           Sobel &                                   95.97 $\pm$ 0.60 &                                   73.43 $\pm$ 0.78 &                                   70.56 $\pm$ 5.56 &                                   96.63 $\pm$ 0.37 \\
                                          &
           BagNet &                                   \textbf{97.26 $\pm$ 0.60} &                                   35.44 $\pm$ 0.80 &                                   41.67 $\pm$ 6.67 &                                   96.30 $\pm$ 0.40 \\
    \midrule \multirow{2}{*}{Co-training} & \begin{tabular}{c}Canny \\
    +BagNet\end{tabular} &                                   96.94 $\pm$ 0.56 &
    \textbf{86.69 $\pm$ 0.56} &                                   \textbf{79.44
    $\pm$ 5.00} &                                   97.53 $\pm$ 0.31 \\
                                          & \begin{tabular}{c}Sobel \\
                                          +BagNet\end{tabular} &
                                          96.81 $\pm$ 0.56 &
                                          84.41 $\pm$ 0.63 &
                                          \textbf{79.44 $\pm$ 5.00} &
                                          \textbf{97.89 $\pm$ 0.29} \\
    \bottomrule
    \end{tabular}
    

    \end{center}
    \vspace{1em}
    \caption{Accuracy of predicting gender on different subpopulations of the
        CelebA dataset.
        We show the accuracy of standard models trained on the pseudo-labels
        produced by different self- or co-training schemes.
        Recall that in the training set all females are blond and all males are
        non-blond (while the unlabeled dataset is balanced).
        It is thus interesting to consider where this correlation
        is reversed.
        We observe that, in these cases, both the standard and BagNet models
        perform quite poorly, even after being self-trained on the unlabeled
        dataset where this correlation is absent.
        At the same time, co-training steers the models away from this
        correlation, resulting in improved performance.
    95\% confidence intervals computed via bootstrap are shown.}
    \label{tab:splitceleba}
\end{table}

\clearpage
\subsection{What if the unlabeled data also contained the spurious correlation?}
\label{app:fullskew}
In Section \ref{sec:spurious}, we assume that the unlabeled data does not contain the spurious correlation present in the labeled data. 
This is often the case when unlabeled data can be collected through a more diverse process than labeled data (for example, by scraping the web 
large scales or by passively collecting data during deployment). This assumption is important: in order to successfully steer models away 
from the spurious correlation during co-training, the process needs to surface examples which contradict the spurious correlation. However, 
if the unlabeled data is also heavily skewed, such examples might be rare or non-existent.

What happens if the unlabeled data is as heavily skewed as the labeled data? We return the setting of a spurious association between 
hair color and gender in CelebA. However, unlike in Section \ref{sec:spurious}, we use an unlabeled dataset that also perfectly correlates 
hair color and gender -- it contains 2000 non-blond males and 2000 blond females. The unlabeled data thus has the same distribution as the labeled
data, and contains no examples that reject the spurious correlation (blond males or non-blond females).

\begin{table}[!h]
    \begin{center}
        \begin{tabular}{llccc}
    \toprule
                                  Methods & Prior(s) &     Labeled Only & \begin{tabular}{c}+Unlabeled \\Self/Co-Training\end{tabular} & \begin{tabular}{c} + Standard model \\with Pseudo-labels \end{tabular} \\
    \midrule
           \multirow{3}{*}{Self-training} & Standard & 67.07 $\pm$ 0.57 &     73.32 $\pm$ 0.55                               &                                   69.13 $\pm$ 0.58 \\
                                          &    Canny & 80.90 $\pm$ 0.49 &     80.47 $\pm$ 0.48                               &                                   76.61 $\pm$ 0.52 \\
                                          &   BagNet & 69.35 $\pm$ 0.55 &     69.21 $\pm$ 0.53                               &                                   71.34 $\pm$ 0.54 \\
    \midrule \multirow{2}{*}{Co-training} &    Canny & 80.90 $\pm$ 0.49 &     \textbf{82.17 $\pm$ 0.47}                               &             \multirow{2}{*}{\textbf{78.53 $\pm$ 0.49}} \\
                                          &  +BagNet & 69.35 $\pm$ 0.55 &     76.52 $\pm$ 0.50                              &                 \\  
    \bottomrule
    \end{tabular}
    \end{center}
    \vspace{1em}
    \caption{Performance of Self-Training and Co-Training techniques when the unlabeled data also contains a complete skew toward hair color (as in the labeled data).
    95\% confidence intervals computed via bootstrap are shown.}
    \label{tab:fullskew}
\end{table}

\paragraph{Self-Training:} Since the unlabeled data follows the spurious correlation between hair color and gender, the standard and 
BagNet models almost perfectly pseudo-label the unlabeled data. Thus, they are simply increasing the number of examples in the training dataset 
but maintaining the same overall distribution. Self-training thus does not change the accuracy for models with these priors significantly. 
In contrast, in the setting in Section \ref{sec:spurious}, there were examples in the unlabeled data which did not align with the spurious correlation 
(blond males and non-blond females). Since they relied mostly on hair color, the standard and BagNet models actively mislabeled these examples 
(i.e, by labeling a blond male as female). Training on these erroneous pseudo-labels actively suppressed any features that were not hair color, causing the standard and Bagnet 
models to perform worse after self-training. 

\paragraph{Co-Training:} In contrast, when performing co-training with the Canny and BagNet priors, the Canny model (which cannot detect hair color) 
will make mistakes on the unlabeled dataset. These mistakes help are inconsistent with a reliance on hair color: due to this regularization, the 
BagNet's accuracy improves from  69.35\% to 76.52\%. Overall, though the gain is not as significant as the setting with a balanced unlabeled dataset, the 
Canny + BagNet co-trained model can mitigate the pitfalls of the BagNet's reliance on hair color and outperform even the canny self-trained model.

\end{document}